\documentclass[10pt,journal,compsoc]{IEEEtran}

\usepackage{amsmath,amssymb,amsfonts}
\usepackage{algpseudocode}
\usepackage{algorithmicx,algorithm}
\usepackage{algpseudocode}
\usepackage{amsmath}
\usepackage{stfloats}
\usepackage{graphicx}
\usepackage{booktabs}
\usepackage{textcomp}
\usepackage{xcolor}
\usepackage{bm}
\usepackage{subfigure}
\usepackage{authblk}
\usepackage{array}
\usepackage{diagbox}
\usepackage{multirow}

\newcolumntype{C}[1]{>{\centering}p{#1}}
\begin{document}


\title{Collaborative Route Planning of UAVs, Workers and Cars for Crowdsensing in Disaster Response}

\author{Lei~Han, Chunyu~Tu, Zhiwen~Yu,~\IEEEmembership{Senior~Member~IEEE}, Zhiyong~Yu,~\IEEEmembership{Member~IEEE}, Weihua~Shan, Liang~Wang, and~Bin~Guo
	\thanks{Lei Han, Zhiwen Yu (corresponding author), Liang Wang and Bin Guo are with School of Computer Science, Northwestern Polytechnical University, Xi’an 710072, China. E-mail: \{leihan, zhiwenyu, liangwang, binguo\}@nwpu.edu.cn.}
	\thanks{Chunyu Tu and Zhiyong Yu are with College of Computer and Data Science, Fuzhou University, Fuzhou 350108, China. E-mail: \{chunyutu, yuzhiyong\}@fzu.edu.cn.}
	\thanks{Weihua Shan is with Innovation Lab, Huawei Cloud Computing Technologies Co.,Ltd, Xi'an 710072, China. E-mail: {ShanWeihua}@huawei.com.}
	\thanks{Manuscript received XX XX, XX; revised XX XX, XX.}}

\markboth{Journal of \LaTeX\ Class Files,~Vol.~14, No.~8, August~2015}%
{Shell \MakeLowercase{\textit{et al.}}: Bare Advanced Demo of IEEEtran.cls for IEEE Computer Society Journals}

\IEEEtitleabstractindextext{%
\begin{abstract}
Efficiently obtaining the up-to-date information in the disaster-stricken area is the key to successful disaster response. Unmanned aerial vehicles (UAVs), workers and cars can collaborate to accomplish sensing tasks, such as data collection, in disaster-stricken areas. In this paper, we explicitly address the route planning for a group of agents, including UAVs, workers, and cars, with the goal of maximizing the task completion rate. We propose MANF-RL-RP, a heterogeneous multi-agent route planning algorithm that incorporates several efficient designs, including global-local dual information processing and a tailored model structure for heterogeneous multi-agent systems. Global-local dual information processing encompasses the extraction and dissemination of spatial features from global information, as well as the partitioning and filtering of local information from individual agents. Regarding the construction of the model structure for heterogeneous multi-agent, we perform the following work. We design the same data structure to represent the states of different agents, prove the Markovian property of the decision-making process of agents to simplify the model structure, and also design a reasonable reward function to train the model. Finally, we conducted detailed experiments based on the rich simulation data. In comparison to the baseline algorithms, namely Greedy-SC-RP and MANF-DNN-RP, MANF-RL-RP has exhibited a significant improvement in terms of task completion rate.

\end{abstract}

}

\maketitle

\section{Introduction}

\begin{figure}[h]
	\centering
	\begin{minipage}[t]{9cm}
		\setlength{\abovecaptionskip}{0cm}   
		\setlength{\belowcaptionskip}{0cm}   
		\centering 
		\includegraphics[width=8.5cm]{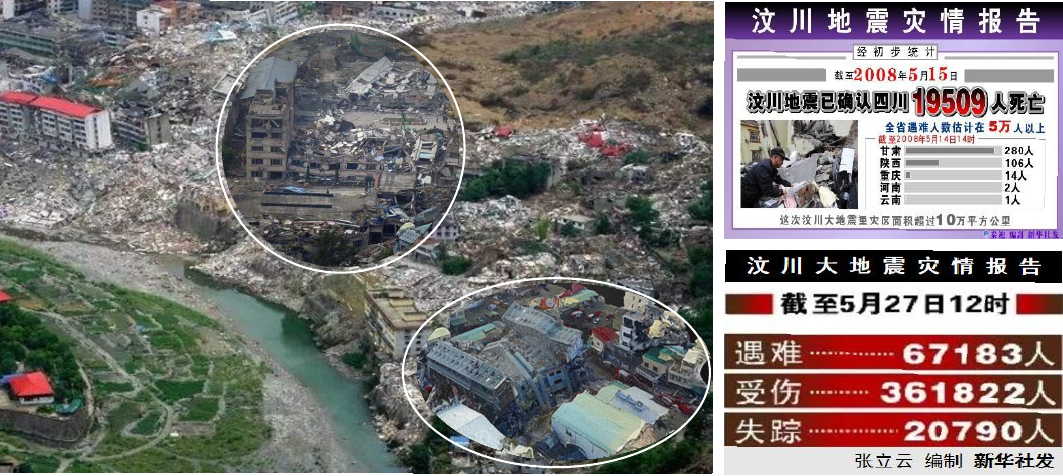}
		\caption{The Wenchuan earthquake, which resulted in a devastating toll: 67,183 deaths, 361,822 injuries, and 20,790 missing persons by 12:00 on May 27, 2008.}
		\label{figure1}
	\end{minipage}
\end{figure}

Devastating disasters, as depicted in Figure \ref{figure1} (e.g., earthquakes), can result in significant loss of life and widespread casualties within a short period of time. In particular, the chances of survival for individuals decrease significantly as the rescue time prolongs. For example, based on the common knowledge of earthquake relief \cite{gode72}, after an earthquake occurs, the survival probability of survivors is approximately 90\% on the first day, but it decreases significantly to around 50\%-60\% on the second day. In such emergencies, rescuers require timely access to the latest information in the disaster-stricken area, as it serves as the foundation for subsequent effective rescue operations.

At present, mobile crowdsensing (MCS) \cite{2011Mobile} is an effective sensing paradigm, which has been widely used in environmental monitoring \cite{2009Common}, public safety \cite{2011Discovery}, intelligent transportation \cite{None2014How} and other fields. However, when a devastating disaster occurs, the environment in the disaster-stricken area becomes extremely complex and dangerous, which greatly limits the mobility of participants. Furthermore, since the traditional MCS relies on the participants and their mobile devices as the basic sensing unit, it is hard to work in the disaster-stricken area that require high sensing accuracy and specific sensing capabilities. With the popularization of unmanned aerial vehicles (UAVs) in recent years, UAVs play a crucial role in disaster response. UAVs, with their capabilities of rapid deployment, high mobility, and the ability to carry high-sensing sensors, can make up for the limitations of traditional MCS. Therefore, many researchers study how to apply UAVs to disaster response \cite{zhou2018mobile, liu2019energy, wang2021energy, liu2020curiosity, dai2022aoi, liu2020energy, liu2019distributed}.

However, the existing researches have two unrealistic assumptions regarding UAVs, which hinder their practical application in disaster-stricken areas. (1) Existing researches assume that UAVs can perform sensing tasks (e.g., data collection) autonomously in the disaster-stricken area. However, the low-altitude environment of disaster-stricken areas poses numerous safety concerns, and many sensing tasks require precise maneuvering of UAVs in this challenging environment.During the execution of sensing tasks, UAVs not only have to navigate around obstacles effectively but also need to accurately detect crucial areas. Without the skilled intervention of professional personnel, it becomes extremely challenging for UAVs to autonomously carry out these sensing tasks. (2) Existing research assumes that UAVs have the capability to autonomously navigate to charging stations for recharging. However, the availability of charging stations specifically designed for UAVs is currently limited in urban areas. Moreover, after a devastating disaster, some charging stations may be damaged or rendered inoperable. Additionally, self-charging for UAVs in outdoor environments without human assistance is extremely challenging. Furthermore, the process of recharging UAVs is time-consuming, which can significantly reduce their operational efficiency. To ensure uninterrupted performance of sensing tasks, a more efficient approach is for cars to directly replace UAV batteries instead of wasting time on UAV charging.

\begin{figure}[h]
	\centering
	\begin{minipage}[t]{9cm}
		\setlength{\abovecaptionskip}{0cm}   
		\setlength{\belowcaptionskip}{0cm}   
		\centering 
		\includegraphics[width=9cm]{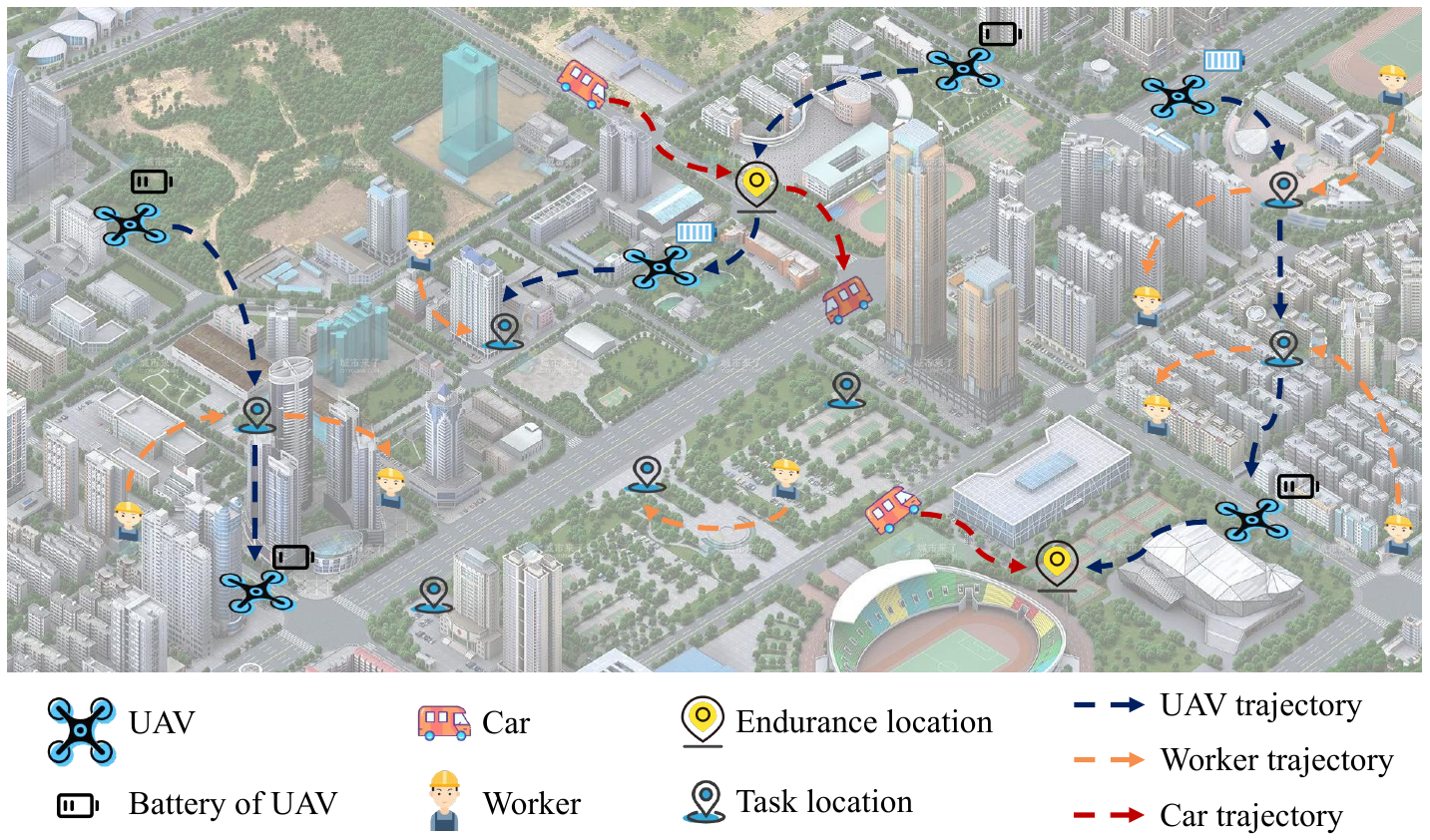}
		\caption{Collaborative route planning of UAVs, workers and cars for crowdsensing.}
		\label{figure2}
	\end{minipage}
\end{figure}

To address the aforementioned challenges, this paper focuses on investigating collaborative route planning for UAVs, workers, and cars to efficiently accomplish sensing tasks, as illustrated in Figure \ref{figure2}. The workers are responsible for the precise manipulation of UAVs at the sensing task locations, where both cars and UAVs can access and where the sensing tasks are located. This is to overcome the limitations of UAVs in autonomous low-altitude maneuvering. Cars, on the other hand, can swiftly replace the batteries of UAVs at designated endurance locations, which are accessible to both cars and UAVs. This enables efficient replenishment of battery power for UAVs. In this particular application scenario, there are two questions that need to be addressed. (1) Why doesn't the workers carry UAVs to perform the sensing tasks by themselves, but move independently and meet UAVs at the sensing task locations? The workers have limited mobility in disaster-stricken areas, and carrying UAVs would further restrict their mobility. Moreover, if the workers carry the UAVs, it would hinder the UAVs' ability to swiftly move between multiple sensing task locations, thus affecting their overall efficiency. (2) Why are UAVs able to autonomously fly between multiple sensing task locations? Unlike the complex low-altitude environment where sensing tasks are performed, UAVs can navigate between multiple locations in the much simpler high-altitude environment. The high-altitude environment poses fewer challenges and obstacles for UAV flight. Additionally, when UAVs are flying between these sensing task locations at higher altitudes, they do not require precise manipulation by professional workers. Furthermore, in our specific application scenario, workers are tasked with manipulating UAVs for sensing tasks, while cars are responsible for replacing UAV batteries. (3) Why can't workers themselves replace the batteries for the UAVs? Due to the limited mobility of workers in disaster-stricken areas, it would be impractical for them to carry spare batteries and hinder their mobility even further. As a result, workers are unable to fulfill the role of replacing UAV batteries.

In recent years, reinforcement learning (RL) has achieved outstanding performance in solving sequential decision-making problems \cite{silver2017mastering, mnih2015human}. Therefore, this paper proposes a collaborative route planning approach for UAVs, workers, and cars in disaster response using multi-agent reinforcement learning (MARL). It is worth noting that in MARL, one agent controls either a single UAV, a worker, or a car. For example, if we need to plan the routes of 5 UAVs, 10 workers, and 3 cars, we would require a total of 18 agents to achieve our goal. However, there are three challenges in using MARL to collaboratively plan the routes of UAVs, workers, and cars for executing sensing tasks. (1) Traditional spatial crowdsourcing only involves two-dimensional matching of users and locations \cite{zhao2016spatial, tong2016online}. 3D spatial crowdsourcing involves three-dimensional matching of users, workers and locations \cite{li2021simultaneous, li2019three}. Our problem involves four-dimensional matching of UAVs, workers, cars and location, which is more complex. (2) The attributes of UAVs, workers and cars vary greatly, such as mobility, endurance and function. In order to expedite the completion of sensing tasks, the route planning for UAVs, workers, and cars necessitates not only efficient spatio-temporal coordination but also appropriate functional alignment. (3) A large parameter scale is not conducive to model convergence. UAVs, workers, and cars exhibit heterogeneity, resulting in different state representations. Therefore, it is necessary to employ a shared neural network for all agents (i.e., UAVs, workers, and cars) to reduce the model's parameter scale. In summary, this work makes the following contributions:

(1) To the best of our knowledge, this work is the first research addressing the collaborative route planning of UAVs, workers, and cars in order to efficiently accomplish sensing tasks for crowdsensing in disaster response. In addition, we prove that the problem is NP-Hard.

(2) To tackle the aforementioned challenges, we propose a MARL-based heterogeneous multi-agent route planning algorithm called MANF-RL-RP. The algorithm comprises two main components, as outlined below.

(a) Global-local dual information processing. First, we mine the spatial features of global information based on convolutional neural networks (CNN) and share them with all agents to reduce the model training cost. Then, we divide the local information of agents into two parts: state information and filtering information. State information is used to guide the agents to make sequential decision. Filtering information is used to filter the non-optional actions to address the issue of sparse rewards in the sequential decision-marking process.

(b) Model structure for heterogeneous multi-agent. We fill in the missing information of workers and cars to use the same data structure to represent the state of UAVs, workers, and cars, then share the same neural network parameter to reduce model parameter scale. Furthermore, we design a reasonable reward function and prove that UAVs, workers, and cars have cooperative relationships, which can guide model training well. Finally, we prove that the sequential decision-making process of agents has the Markov property, which simplifies the agent network structure.

(3) We conducted detailed experiments based on the rich simulation data. In comparison to the baseline algorithms, namely Greedy-SC-RP and MANF-DNN-RP, MANF-RL-RP has exhibited a significant improvement in terms of task completion rate. The experimental code and data examples for this paper can be referenced from \cite{codeanddata}.

The contents of this paper are arranged as follows: Section \ref{section2} discusses some related works; Section \ref{section3} formulates the collaborative route planning of workers, cars and UAVs for crowdsensing in disaster response; Section \ref{section4} models some concepts of the sequential decision-making process in this problem, and Section \ref{section5} implements the heterogeneous multi-agent route planning algorithms MANF-DNN-RP and MANF-RL-RP based on the concepts modeled in Section \ref{section4}; then the experiments are conducted in Section \ref{section6}; finally, conclusions and future work are summarized in Section \ref{section7}.

\section{Related Work}
\label{section2}
\subsection{Task allocation of traditional MCS}

Task assignment of traditional Mobile Crowd Sensing (MCS) can be divided into two categories: single-task assignment and multi-tasks assignment. Single-task assignment focuses on the relationship between the spatial-temporal coverage of tasks and limited sensing resources, such as limited participants or sensing budget. For example, under the fixed sensing resources, maximize data quality or the overall utility of system \cite{reddy2009using, zhang2014crowdrecruiter, xiong2015icrowd, song2014qoi}. Alternatively, minimize sensing cost or the number of participants while ensuring data quality \cite{karaliopoulos2015user, yu2018participant, wang2017multi}. From single-task assignment to multi-tasks assignment, we need to take into account the following issues. From an optimization perspective, one must consider how to balance the quality of sensing data for multiple tasks while ensuring the quality of sensing data for each individual task \cite{zhang2015quality, wang2018multi}. From a temporal perspective, the duration of different tasks may vary. It is necessary to consider corresponding task allocation strategies to address the varying time scales of different tasks \cite{li2015dynamic}. From a spatial perspective, the spatial granularity of different tasks may also vary, and there may be inclusion relationships among them. It is necessary to address the spatial overlap between different tasks \cite{wang2018heterogeneous}. From the perspective of sensing content, different tasks may involve the same data. By assigning tasks based on data attributes, we can avoid data redundancy \cite{wang2019user, han2021online}. In traditional MCS, participants and their mobile devices are considered as the fundamental sensing units. However, the mobility of participants and the sensing capabilities of mobile devices are limited.

\subsection{UAVs for MCS}

In response to the limitations of traditional MCS, researchers have explored the integration of Unmanned Aerial Vehicles (UAVs) in MCS. UAVs possess exceptional maneuverability and can be equipped with capabilities of rapid deployment, high mobility, and the ability to carry high-sensing sensors. For example, UAVs can act as aerial base stations to assist in data transmission. Liu et al. considered to use a group of UAVs as aerial base stations to move around and collect data from multiple MCS users \cite{dai2022aoi}. Liu et al. studied how to tackle the problem that a group of UAVs energy-efficiently and cooperatively collect data from low-level sensors, while charging the battery from multiple randomly deployed charging stations \cite{liu2020energy}. Liu et al. designed a fully-distributed control solution to navigate a group of UAVs, as the mobile base stations to fly around a target area, to provide long-term communication coverage for the ground mobile users \cite{liu2019distributed}. In addition, UAVs can also be used to collect data. Zhou et al. considered the fixed-wing UAV-aided MCS system and investigate the corresponding joint route planning and task assignment problem from an energy efficiency perspective \cite{zhou2018mobile}. Liu et al. navigated a group of UAVs to move around a target area to maximize their total amount of collected data with the limited energy reserve, while geographical fairness among those point-of-interests should also be maximized \cite{liu2019energy}. Liu et al. explicitly considered to navigate a group of UAVs in a 3-dimensional disaster work zone to maximize the amount of collected data, geographical fairness, energy efficiency, while minimizing data dropout due to limited transmission rate \cite{wang2021energy}. Liu et al. deployed UAVs in remote or hazardous areas to carry on long-term and hash tasks to achieve an optimal trade-off between maximizing the collected amount of data and coverage fairness, and minimizing the overall energy consumption of workers \cite{liu2020curiosity}. However, existing researches have made two unrealistic assumptions regarding UAVs, making it challenging for UAVs to be effectively utilized in disaster-stricken areas. Firstly, the assumption that UAVs can autonomously perform sensing tasks is not practical. Secondly, the assumption that UAVs can independently go to charging stations to recharge themselves is also unrealistic. In light of these challenges, we focus on studying collaborative route planning for UAVs, workers, and cars to address these issues. Workers are responsible for precise manipulation of UAVs at task locations, while cars play a crucial role in replacing UAV batteries to ensure sufficient battery life.

\subsection{MARL for cooperative tasks}

In recent years, RL has achieved outstanding performance in solving sequential decision-making problems \cite{silver2017mastering, mnih2015human}. In the research problem of this paper, the collaborative execution of sensing tasks requires the cooperation of multiple agents. As the number of agents increases, the dimension of joint actions grows exponentially \cite{2016TaskMe, 2017Joint, 2017Multiuser}. Treating multiple agents as a single agent based on joint actions is not a feasible approach. Therefore, the problem can only be solved using MARL (Multi-Agent Reinforcement Learning). Common MARL algorithms for cooperative tasks include MADDPG \cite{lowe2017multi}, COMA \cite{foerster2018counterfactual}, VDN \cite{sunehag2017value}, QMIX \cite{rashid2018qmix}, WQMIX \cite{rashid2020weighted} and QTRAN \cite{son2019qtran}, etc. MADDPG and COMA are implemented based on Actor-Critic framework. In MADDPG, each agent is associated with an actor and a critic, and each critic observes the states and actions of all agents. However, when the number of agents is large, the model becomes excessively large. On the other hand, COMA has a centralized critic and distributed actors. The centralized critic guides the training of distributed actors. However, when the number of agents is large, the state input to the critic becomes extensive, resulting in difficulties in the convergence of the critic network. VDN, QMIX, WQMIX, and QTRAN are value decomposition-based algorithms. In these algorithms, each agent has its own utility function, and the central action-value function is a combination of the utility functions of all agents. VDN assumes a linear sum relationship between the central action-value function and the utility functions, which limits the representation capability of the action-value function. To address this limitation, QMIX introduces a non-linear relationship between the action-value function and the utility functions. In terms of implementation, QMIX incorporates a mixing module that combines the utility functions of the agents, significantly enhancing the expressiveness of the action-value function. While WQMIX guarantees the satisfaction of monotonicity constraint in the output action-value functions of QMIX, and QTRAN provides necessary and sufficient conditions for decomposability of QMIX's action-value functions, their practical performance is hindered by numerous constraints and assumptions. Therefore, our proposed algorithm is based on QMIX for its practical applicability.

\section{Problem definition}
\label{section3}

In this section, we begin by providing definitions for key concepts and subsequently formulate the collaborative route planning of  UAVs, workers and cars for crowdsensing in disaster response.

{\itshape Definition 1.} Discrete area set $AREA = \{ are{a_0},...,are{a_m},...\}$. $are{a_m} = \left\langle {obs{t_m},tas{k_m}} \right\rangle$ represents the $m-th$ area in $AREA$. $obs{t_m}$ is the obstacle identification of $are{a_m}$. When there are an obstacle in $are{a_m}$, $obs{t_m}$ is marked as 1, otherwise 0. $tas{k_m}$ is the sensing task identification of $are{a_m}$. When there are a sensing task in $are{a_m}$, $tas{k_m}$ is marked as 1, otherwise 0.

%

{\itshape Definition 2.} UAVs set $UAV = \{ ua{v_0},...,ua{v_i},...\}$. $ua{v_i} = \left\langle {uLo{c_i^t},uRg{e_i^t},uPo{w_i^t},uCs{p_i}} \right\rangle$ represents the $i-th$ UAV in $UAV$. $uLoc_i^t \in AREA$ represents the location of $ua{v_i}$ at moment $t,(t \ge 0)$. $uRge_i^t \in [\emptyset ,AREA)$ represents the range that $ua{v_i}$ can move within the time step $[t,t + 1)$. $uPow_i^t \in [0,1]$ represents the remaining power of $ua{v_i}$ at the moment $t$. $uCs{p_i}$ represents the power consumed by $ua{v_i}$ in a time step. We assume that $ua{v_i}$ consumes the same power at any time step.

{\itshape Definition 3.} Workers set $Worker = \{ wk{r_0},...,wk{r_j},...\}$. $wk{r_j} = \left\langle {wLo{c_j^t},wRg{e_j^t}} \right\rangle$ represents the $j-th$ worker in $Worker$. $wLoc_j^t \in AREA$ represents the location of $wk{r_j}$ at moment $t,(t \ge 0)$. $wRge_j^t \in [\emptyset ,AREA)$ represents the range that $wk{r_j}$ can move within the time step $[t,t + 1)$.

{\itshape Definition 4.} Cars set $Car = \{ ca{r_0},...,ca{r_k},...\}$. $ca{r_k} = \left\langle {cLo{c_k^t},cRg{e_k^t}} \right\rangle$ represents the $k-th$ car in $Car$. $cLoc_k^t \in AREA$ represents the location of $ca{r_k}$ at moment $t,(t \ge 0)$. $cRge_k^t \in [\emptyset ,AREA)$ represents the range that $ca{r_k}$ can move within the time step $[t,t + 1)$.

In real-world scenarios, UAVs, workers, and cars exhibit variations in mobility. To formulate this problem more clearly, we make the following assumption. If $obs{t_m} = 1$, neither UAVs, workers nor cars can reach $are{a_m}$, otherwise there is no restriction, refer to \cite{liu2019energy, wang2021energy, liu2020multi, liu2019distributed}. In addition, workers and UAVs can perform the sensing task when they meet at the sensing task location. The cars can replace the battery of UAVs when the UAVs meet the cars at designated endurance locations. In this paper, we assert that any unobstructed location can function as an endurance locations.

{\itshape Definition 5.} The routes set of UAVs, $UTRA = \{ uTr{a_0},...,uTr{a_i},...\}$, $uTr{a_i} = \{ uLoc_i^0,...,uLoc_i^t\}$ represents the route of $ua{v_i}$.

{\itshape Definition 6.} The routes set of Workers, $WTRA = \{ wTr{a_0},...,wTr{a_j},...\}$, $wTr{a_j} = \{ wLoc_j^0,...,wLoc_j^t\}$ represents the route of $wk{r_j}$.

{\itshape Definition 7.} The routes set of Cars, $CTRA = \{ cTr{a_0},...,cTr{a_k},...\}$, $cTr{a_k} = \{ cLoc_k^0,...,cLoc_k^t\}$ represents the route of $ca{r_k}$.

Before defining the problem of this paper, we need to be introduce the following constraints.

(1) When UAVs' battery is low, UAVs will stop moving, see Equation (\ref{eq3}).

\begin{small}
	\begin{equation}
	uRge_i^t = \emptyset ,\ {\rm{if}}\ uPow_i^t < uCs{p_i}
	\label{eq3}
	\end{equation}
\end{small}

(2) UAVs, workers, and cars cannot move to obstacles locations, see Equation (\ref{eq4}).

\begin{small}
	\begin{equation}
	uLoc_i^{t}.obs{t_m} \ne 1\ \& \ wLoc_j^{t}.obs{t_m} \ne 1\ \& \ cLoc_k^{t}.obs{t_m} \ne 1
	\label{eq4}
	\end{equation}
\end{small}

(3) If UAVs meet the workers at a sensing task locations, the sensing task will be performed, see Equation (\ref{eq5}).

\begin{small}
	\begin{equation}
	uLoc_i^t.tas{k_m} = 0,\ {\rm{if}}\ uLoc_i^t = wLoc_j^t\ \& \ uLoc_i^t.tas{k_m} = 1
	\label{eq5}
	\end{equation}
\end{small}

(4) UAVs, workers and cars cannot move beyond the movable range within the time step $[t,t + 1)$, see Equation (\ref{eq6}).

\begin{small}
	\begin{equation}
	uLoc_i^{t + 1} \in uRge_i^t\ \& \ wLoc_j^{t + 1} \in wRge_j^t\ \& \ cLoc_k^{t + 1} \in cRge_k^t
	\label{eq6}
	\end{equation}
\end{small}

(5) If UAVs meet the cars, replace UAVs' battery. Otherwise, the power of UAVs will reduce or be unchanged, see Equation (\ref{eq7}).

\begin{small}
	\begin{equation}
	uPow_i^{t + 1} = \left\{ {\begin{array}{*{20}{c}}
		{1,\ {\rm{if}}\ uLoc_i^{t + 1} = cLoc_k^{t + 1}} \qquad\ \qquad\ \qquad\ \ \ \ \, \vspace{3pt} \\
		\begin{array}{l}
		uPow_i^t - uCs{p_i}, \vspace{3pt} \\
		\quad\ {\rm{if}}\ uLoc_i^{t + 1} \ne cLoc_k^{t + 1}\& uPow_i^t \ge uCs{p_i},
		\end{array} \vspace{3pt} \\
		\begin{array}{l}
		uPow_i^t \vspace{3pt} \\
		\quad\ {\rm{if}}\ uLoc_i^{t + 1} \ne cLoc_k^{t + 1}\& uPow_i^t < uCs{p_i}
		\end{array}
		\end{array}} \right.
	\label{eq7}
	\end{equation}
\end{small}

\textbf{Problem 1} (collaborative route planning of UAVs, workers and cars for crowdsensing in disaster response): Given discrete area set $AREA$, UAVs set $UAV$, workers set $Worker$, cars set $Car$, and upper limit of sensing time $TimeLimit$. Determine the routes set of UAVs $UTRA$, the routes set of workers $WTRA$ and the routes set of cars $CTRA$ during $[0,TimeLimit]$ to maximize the sensing tasks completion $\sum\limits_m {task_m^0}  - \sum\limits_m {task_m^{TimeLimit}}$. Formally,

\begin{small}
	\begin{equation}
	\begin{array}{l}
	{\rm{\textbf{confirm}}}\ UTRA,WTRA,CTRA \vspace{3pt} \\
	{\rm{\textbf{max}}}\ \sum\limits_m {task_m^0}  - \sum\limits_m {task_m^{TimeLimit}} \vspace{3pt} \\
	s.t.\  \rm{constraints \ (1),(2),(3),(4),(5)}
	\end{array} 
	\label{eq8}
	\end{equation}
\end{small}

\textbf{Lemma 1.} \textbf{Problem 1} is NP-Hard.

\emph{Proof:} Assume that UAVs can move indefinitely and complete sensing tasks alone without workers and cars. Without the function matching between different agents, \textbf{Problem 1} can be expressed as \textbf{Problem 2}:

\begin{small}
	\begin{equation}
	\begin{array}{l}
	{\rm{\textbf{confirm}}}\ UTRA \vspace{3pt} \\
	{\rm{\textbf{max}}}\ \sum\limits_m {task_m^0}  - \sum\limits_m {task_m^{TimeLimit}}  \vspace{3pt} \\
	s.t.\ uLoc_i^{t}.obs{t_m} \ne 1 \vspace{3pt} \\
	\quad\ \ uLoc_i^t.tas{k_m} = 0,\ {\rm{if}}\ uLoc_i^t.tas{k_m} = 1 \vspace{3pt} \\ 
	\quad\ \ uLoc_i^{t + 1} \in uRge_i^t
	\end{array}
	\label{eq9}
	\end{equation}
\end{small}

\textbf{Problem 2} is a specific instance of \textbf{Problem 1}, thus reducing \textbf{Problem 2} to \textbf{Problem 1}.

Then, we assume that there is only one UAVs in the UAVs set $UAV$, and \textbf{Problem 2} can be expressed as \textbf{Problem 3}:

\begin{small}
	\begin{equation}
	\begin{array}{l}
	{\rm{\textbf{confirm}}}\ uTr{a_0}  \vspace{3pt} \\
	{\rm{\textbf{max}}}\ \sum\limits_m {task_m^0}  - \sum\limits_m {task_m^{TimeLimit}}  \vspace{3pt} \\
	s.t.\ uLoc_0^t.obs{t_m} \ne 1 \vspace{3pt} \\
	\quad\ \ uLoc_0^t.tas{k_m} = 0,\ {\rm{if}}\ uLoc_0^t.tas{k_m} = 1 \vspace{3pt} \\ 
	\quad\ \ uLoc_0^{t + 1} \in uRge_0^t
	\end{array}
	\label{eq10}
	\end{equation}
\end{small}

\textbf{Problem 3} is a specific instance of \textbf{Problem 2}, thus reducing \textbf{Problem 3} to \textbf{Problem 2}.

Because \textbf{Problem 3} is a subset selection problem with time series, it is NP-Hard \cite{nemhauser1978analysis}. Therefore, \textbf{Problem 1} is NP-Hard.

\section{Problem modeling}
\label{section4}

We model \textbf{Problem 1} as a Markov decision process (MDP), defined as a tuple $(\left\langle {S,O} \right\rangle ,A,p,r,\gamma )$.

\subsection{State space}

In this paper, we divide the information into two parts: global information and local information of agents (i.e., UAVs, workers and cars).

\begin{figure}[h]
	\vspace{0cm} 
	\setlength{\abovecaptionskip}{0cm}   
	\setlength{\belowcaptionskip}{0cm}   
	\subfigcapskip=-0.1cm 
	\centering
	\subfigure[$obstDis{t^t}$]{
		\begin{minipage}[t]{0.33\linewidth}
			\centering
			\includegraphics[width=2.7cm]{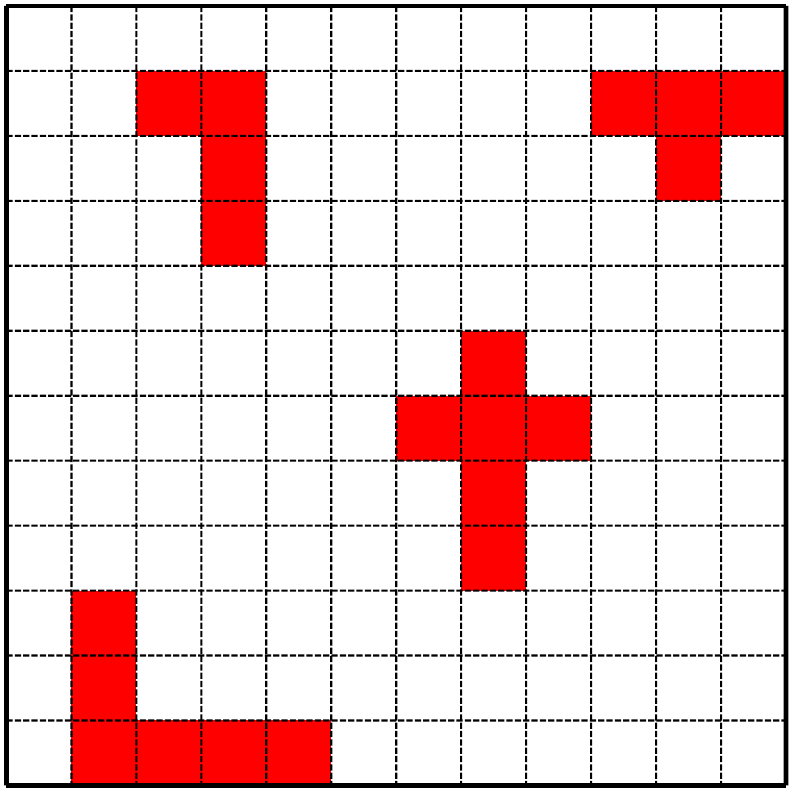}
			\label{figure101}
		\end{minipage}%
	}%
	\subfigure[$taskDis{t^t}$]{
		\begin{minipage}[t]{0.33\linewidth}
			\centering
			\includegraphics[width=2.7cm]{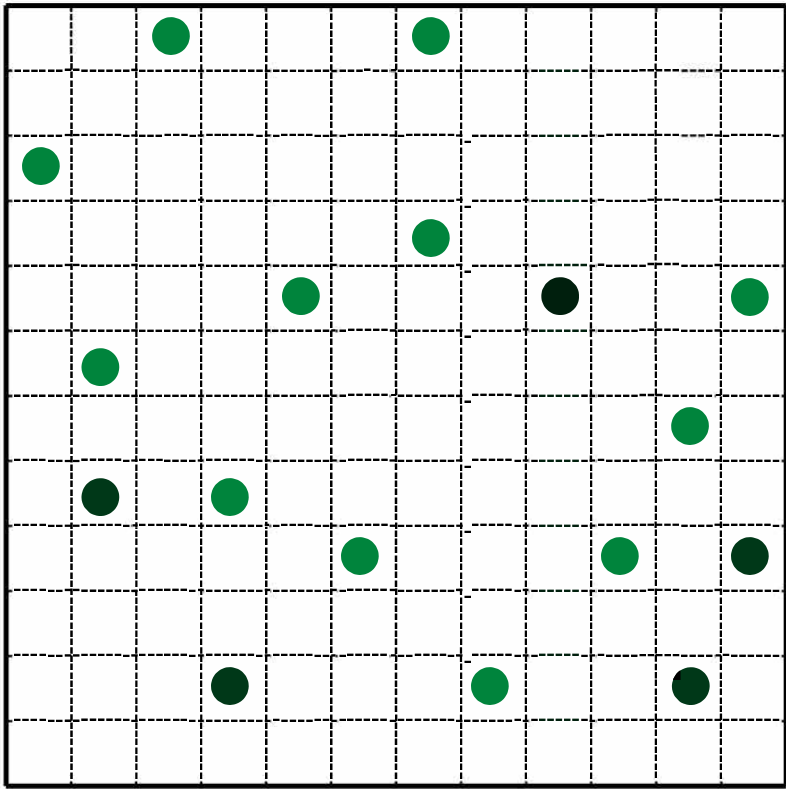}
			\label{figure102}
		\end{minipage}%
	}%
	\subfigure[$urgeDis{t^t}$]{
		\begin{minipage}[t]{0.33\linewidth}
			\centering
			\includegraphics[width=2.7cm]{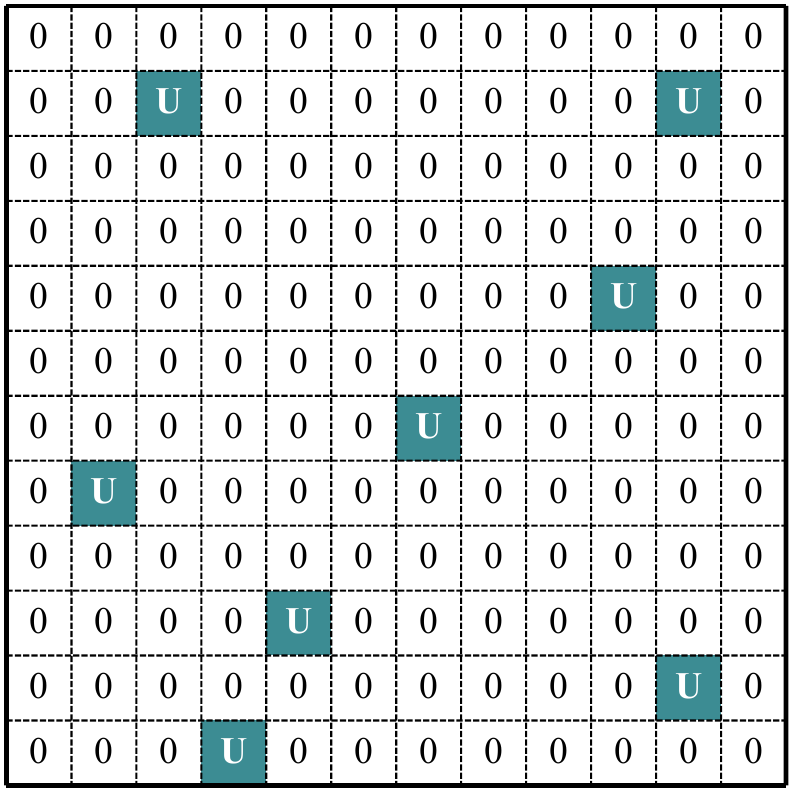}
			\label{figure103}
		\end{minipage}%
	}%

	\subfigure[$workDis{t^t}$]{
		\begin{minipage}[t]{0.33\linewidth}
			\centering
			\includegraphics[width=2.7cm]{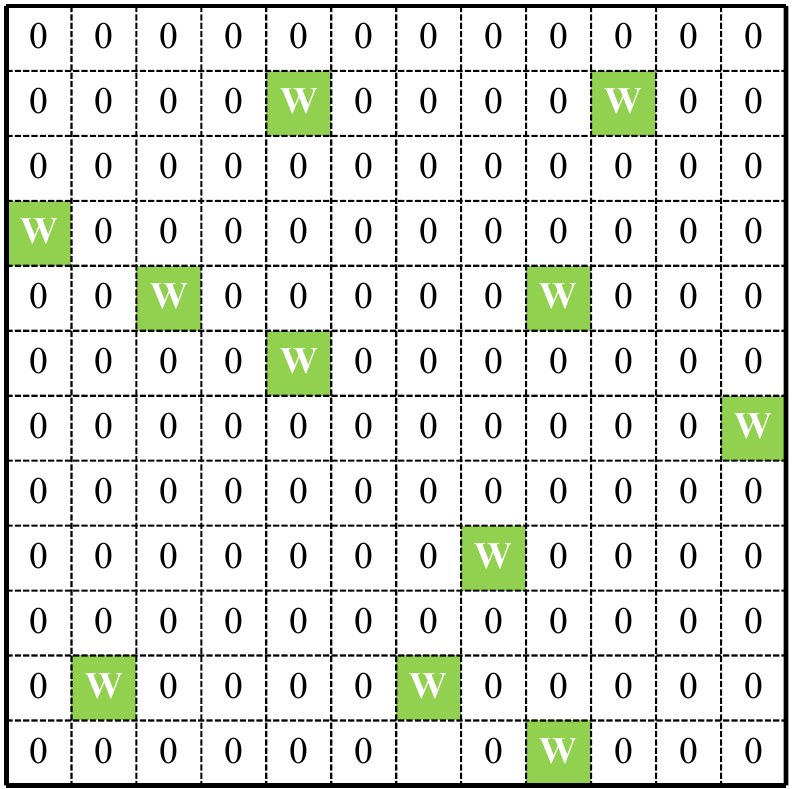}
			\label{figure101}
		\end{minipage}%
	}%
	\subfigure[$carDis{t^t}$]{
		\begin{minipage}[t]{0.33\linewidth}
			\centering
			\includegraphics[width=2.7cm]{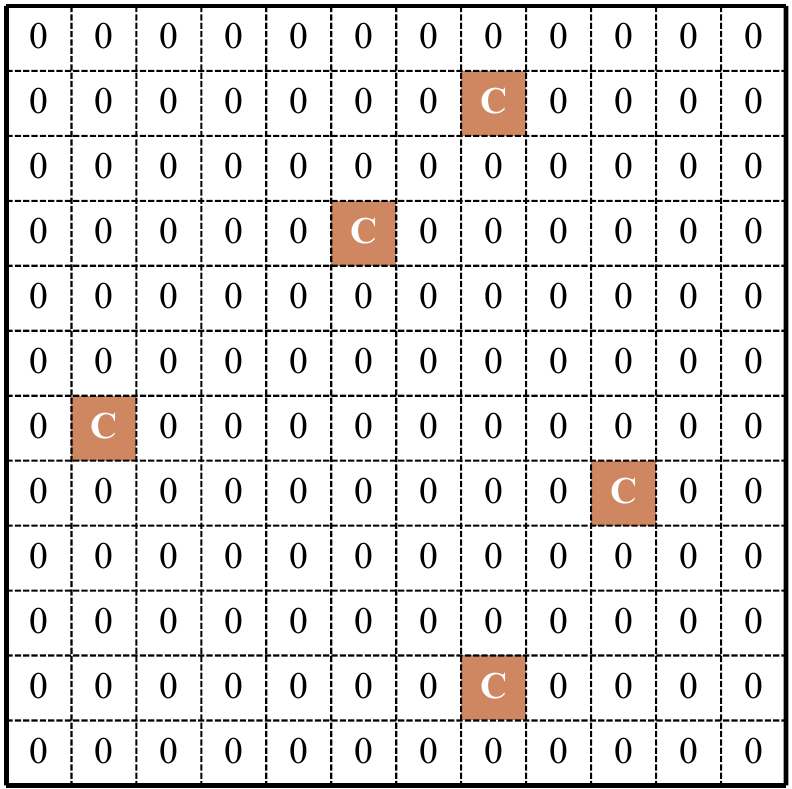}
			\label{figure102}
		\end{minipage}%
	}%
	\centering
	\caption{Global information at the moment $t$.}
	\label{figure3}
\end{figure}

As shown in Figure \ref{figure3}, global information includes $obstDis{t^t}$, $taskDis{t^t}$, $urgeDis{t^t}$, $workDis{t^t}$ and $carDis{t^t}$. $obstDis{t^t}$ represents the location distribution of obstacles at the moment $t$. $taskDis{t^t}$ represents the location distribution of sensing tasks at the moment $t$. $urgeDis{t^t}$ represents the urgency distribution to replace UAVs' battery at the moment $t$, which measures the cumulative urgency of replacing batteries for all UAVs in different locations. $workDis{t^t}$ represents the workers distribution at the moment $t$. $carDis{t^t}$ represents the cars distribution at the moment $t$.

\begin{figure}[h]
	\vspace{0cm} 
	\setlength{\abovecaptionskip}{0cm}   
	\setlength{\belowcaptionskip}{0cm}   
	\subfigcapskip=-0.1cm 
	\centering
	\subfigure[$agentLoc_{ijk}^t$]{
		\begin{minipage}[t]{0.33\linewidth}
			\centering
			\includegraphics[width=2.7cm]{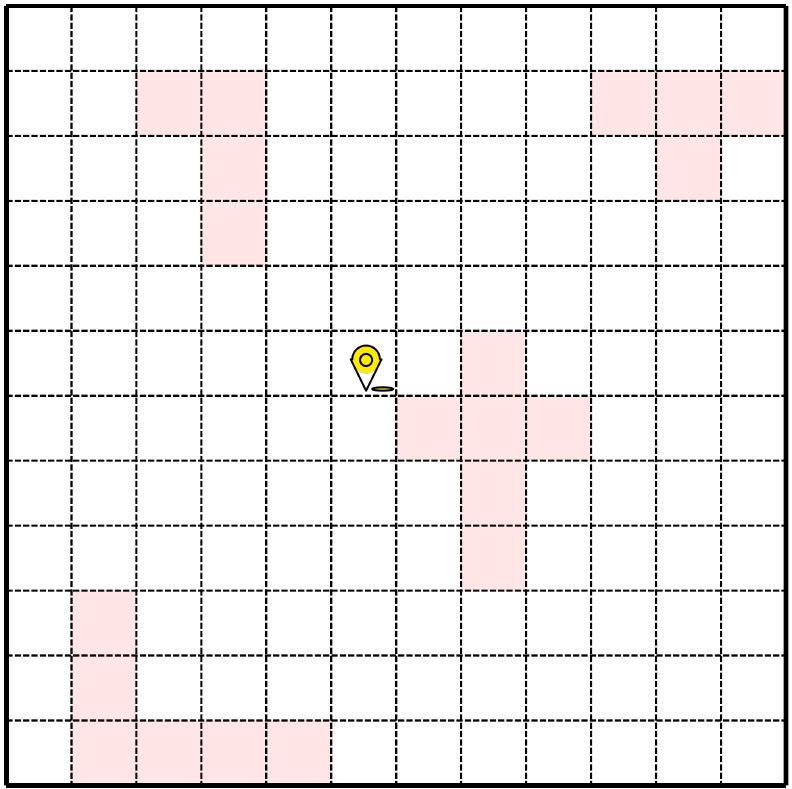}
			\label{figure101}
		\end{minipage}%
	}%
	\subfigure[$agentArriv_{ijk}^t$]{
		\begin{minipage}[t]{0.33\linewidth}
			\centering
			\includegraphics[width=2.7cm]{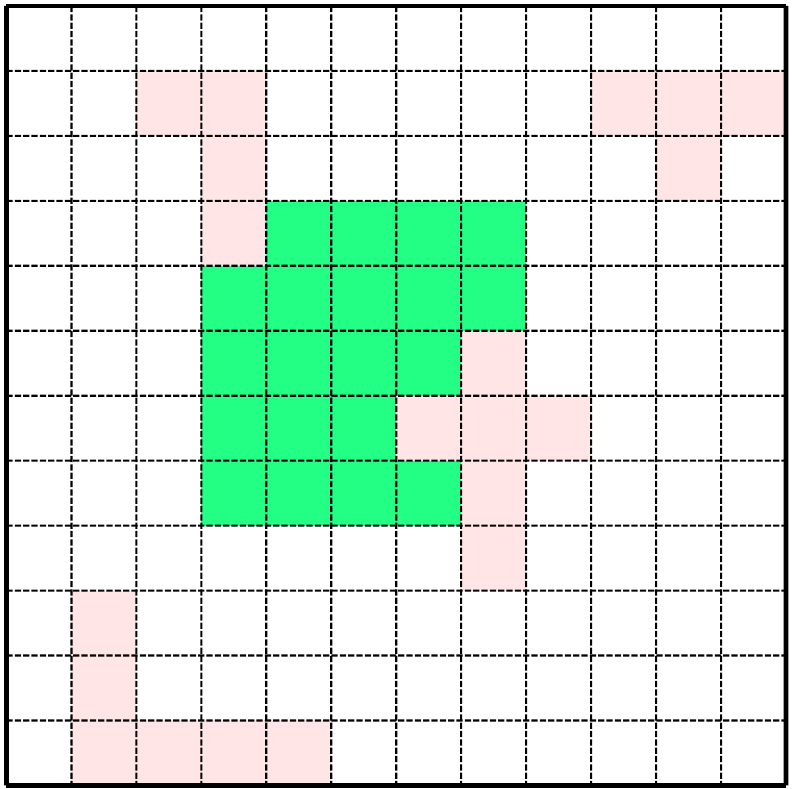}
			\label{figure102}
		\end{minipage}%
	}%
	\subfigure[$urge_{ijk}^t$]{
		\begin{minipage}[t]{0.33\linewidth}
			\centering
			\includegraphics[width=2.4cm]{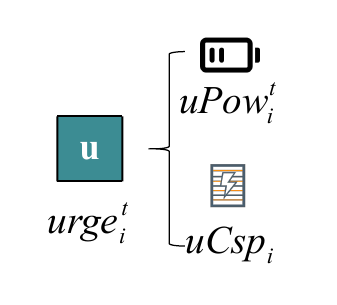}
			\label{figure103}
		\end{minipage}%
	}%
	
	\subfigure[$agentID_{ijk}$]{
		\begin{minipage}[t]{1\linewidth}
			\centering
			\includegraphics[width=9cm]{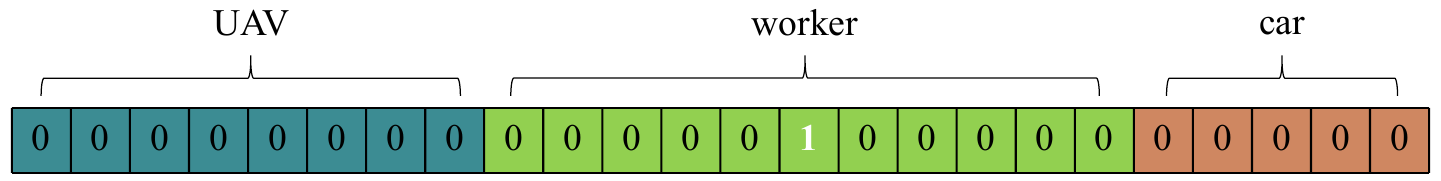}
			\label{figure101}
		\end{minipage}%
	}%
	\centering
	\caption{Local information of the $ijk-th$ agent at the moment $t$.}
	\label{figure4}
\end{figure}

As shown in Figure \ref{figure4}, local information of agents includes $agentLoc_{ijk}^t$, $agentArriv_{ijk}^t$, $urge_{ijk}^t$ and $agentID_{ijk}$, note that $(ijk = i/j/k)$. $agentLoc_{ijk}^t$ represents the location of the $ijk-th$ agent at the moment $t$. $agentArriv_{ijk}^t$ represents the areas that the $ijk-th$ agent can reach within the time step $[t,t + 1)$. In real-world environments, the optional range $agentArriv_{ijk}^t$ of all agents (UAVs, workers and cars) can be predefined based on the actual situation, and the optional range $agentArriv_{ijk}^t$ has already eliminated unreachable locations. $urge_{ijk}^t$ represents the urgency of the $ijk-th$ agent to replace its battery at the moment $t$. $agentID_{ijk}$ represents the ID numbers of the $ijk-th$ agent. $agentID_{ijk}$ is implemented based on one-hot encoding. For example, if the encoding at the ijk-th position of Fig.4 (d) is 1, it represents the ijk-th agent.

In this paper, the evaluation of $urge_{ijk}^t$ needs to meet the following three conditions.

(1) The more the power of the $ijk-th$ agent, the less its urgency. The urgency has a practical physical meaning. When $uav_i$ has infinite power, $uav_i$ would not need to replace its battery, and its urgency should be 0. Finally, we think that workers and cars have infinite power. Therefore, the urgency of workers and cars should be 0.

(2) The larger the power of the $ijk-th$ agent, the less sensitive the urgency of $uav_i$. Therefore, as the power of $uav_i$ increases, the decreasing speed of its urgency becomes small.

(3) The urgency of different agents can be added and the urgency of different areas can be compared. Therefore, there should be an upper limit with the urgency of $uav_i$.

Based on the above three conditions, we define the functional relationship among its current power $uPow_i^t$, power consumption $uCs{p_i}$ in a time step and urgency $urge_i^t$, see Equation (\ref{eq11}). "floor()" means round down function.

\begin{small}
	\begin{equation}
	urge_i^t = \frac{1}{{{e^{{\rm{floor}}({\raise0.7ex\hbox{${uPow_i^t}$} \!\mathord{\left/
							{\vphantom {{uPow_i^t} {uCs{p_i}}}}\right.\kern-\nulldelimiterspace}
						\!\lower0.7ex\hbox{${uCs{p_i}}$}})}}}}
	\label{eq11}
	\end{equation}
\end{small}

So, we can obtain,

\noindent $\begin{array}{l}
\displaystyle \Delta urge_i^t = \frac{{\partial urge_i^t}}{{\partial {\rm{floor}}({\raise0.7ex\hbox{${uPow_i^t}$} \!\mathord{\left/
				{\vphantom {{uPow_i^t} {uCs{p_i}}}}\right.\kern-\nulldelimiterspace}
			\!\lower0.7ex\hbox{${uCs{p_i}}$}})}} = \frac{{\partial \frac{1}{{{e^{{\rm{floor}}({\raise0.7ex\hbox{${uPow_i^t}$} \!\mathord{\left/
								{\vphantom {{uPow_i^t} {uCs{p_i}}}}\right.\kern-\nulldelimiterspace}
							\!\lower0.7ex\hbox{${uCs{p_i}}$}})}}}}}}{{\partial {\rm{floor}}({\raise0.7ex\hbox{${uPow_i^t}$} \!\mathord{\left/
				{\vphantom {{uPow_i^t} {uCs{p_i}}}}\right.\kern-\nulldelimiterspace}
			\!\lower0.7ex\hbox{${uCs{p_i}}$}})}} \vspace{3pt} \\
\displaystyle \qquad \ \ \ \ =  - (\frac{1}{{{e^{{\rm{floor}}({\raise0.7ex\hbox{${uPow_i^t}$} \!\mathord{\left/
						{\vphantom {{uPow_i^t} {uCs{p_i}}}}\right.\kern-\nulldelimiterspace}
					\!\lower0.7ex\hbox{${uCs{p_i}}$}})}}}}) < 0
\end{array}$

Besides, ${\rm{floor}}({\raise0.7ex\hbox{${uPow_i^t}$} \!\mathord{\left/
		{\vphantom {{uPow_i^t} {uCs{p_i}}}}\right.\kern-\nulldelimiterspace}
	\!\lower0.7ex\hbox{${uCs{p_i}}$}}) \propto uPow_i^t$

So, $urge_i^t$ and ${uPow_i^t}$ are inversely proportional. When $uPow_i^t =  + \infty$, $urge_i^t$ is the smallest, which is 0. Condition (1) is satisfied.

Besides, $\displaystyle \begin{array}{l}
\frac{{\partial \Delta urge_i^t}}{{\partial {\rm{floor}}({\raise0.7ex\hbox{${uPow_i^t}$} \!\mathord{\left/
				{\vphantom {{uPow_i^t} {uCs{p_i}}}}\right.\kern-\nulldelimiterspace}
			\!\lower0.7ex\hbox{${uCs{p_i}}$}})}} = \frac{{\partial ( - \frac{1}{{{e^{{\rm{floor}}({\raise0.7ex\hbox{${uPow_i^t}$} \!\mathord{\left/
								{\vphantom {{uPow_i^t} {uCs{p_i}}}}\right.\kern-\nulldelimiterspace}
							\!\lower0.7ex\hbox{${uCs{p_i}}$}})}}}})}}{{\partial {\rm{floor}}({\raise0.7ex\hbox{${uPow_i^t}$} \!\mathord{\left/
				{\vphantom {{uPow_i^t} {uCs{p_i}}}}\right.\kern-\nulldelimiterspace}
			\!\lower0.7ex\hbox{${uCs{p_i}}$}})}} \vspace{3pt} \\
\displaystyle \qquad \qquad \qquad \qquad \quad = \frac{1}{{{e^{{\rm{floor}}({\raise0.7ex\hbox{${uPow_i^t}$} \!\mathord{\left/
						{\vphantom {{uPow_i^t} {uCs{p_i}}}}\right.\kern-\nulldelimiterspace}
					\!\lower0.7ex\hbox{${uCs{p_i}}$}})}}}} > 0
\end{array}$

So, $\Delta urge_i^t$ and ${uPow_i^t}$ are proportional, and $\Delta urge_i^t < 0$. Condition (2) is satisfied.

When $uPow_i^t = 0$, $urge_i^t$ is the largest, which is 1. Condition (3) is satisfied.

Based on the global information and local information of agents, we can construct global state $S = \{ {s^0},...,{s^t},...\}$ and local state of agents $O = \{ \{ o_0^0,...,o_{ijk}^0,...\} ,...,\{ o_0^t,...,o_{ijk}^t,...\} ,...\}$, refer to Equation (\ref{eq111}) and Equation (\ref{eq112}) for details.

\begin{small}
	\begin{equation}
	{s^t} = \{ obstDis{t^t},taskDis{t^t},urgeDis{t^t},workDis{t^t},carDis{t^t}\} 
	\label{eq111}
	\end{equation}
\end{small}

\begin{small}
	\begin{equation}
	o_{ijk}^t = \{ {s^t},agentLoc_{ijk}^t,agentI{D_{ijk}},urge_{ijk}^t\}
	\label{eq112}
	\end{equation}
\end{small}

%

$agentArriv_{ijk}^t$ is used to filter the unreachable locations for the $ijk-th$ agent at the moment $t$.

\subsection{Action space}

Action set $A = \{ \{ a_0^0,...,a_{ijk}^0,...\} ,...,\{ a_0^t,...,a_{ijk}^t,...\} ,...\}$. $a_{ijk}^t = uLoc_i^t/wLoc_j^t/cLoc_k^t$ represents the location that the $ijk-th$ agent will reach within the time step $[t,t + 1)$. Based on the state space, we know that $a_{ijk}^t$ is only affected by $o_{ijk}^t$ and $agentArriv_{ijk}^t$. Therefore, $\{ a_{ijk}^0,...,a_{ijk}^t\}$ are independent with each other.

\subsection{State transition}

$\left\langle {S,O} \right\rangle  \times A \times \left\langle {S,O} \right\rangle  \to p,(p \in [0,1])$ represents the probability distribution of a state transition $p( \{ {s^{t + 1}},o_0^{t + 1},...,o_{ijk}^{t + 1},...\} |\{ {s^t},o_0^t,...,o_{ijk}^t,...\} ,\{ a_0^t,...,a_{ijk}^t,...\} )$, in which the current state is $\{ {s^t},o_0^t,...,o_{ijk}^t,...\}$. When action $\{ a_0^t,...,a_{ijk}^t,...\}$ is chosen, the state is transitioned to a new state $\{ {s^{t + 1}},o_0^{t + 1},...,o_{ijk}^{t + 1},...\}$

\textbf{Lemma 2.} $\{ \{ o_0^0,...,o_{ijk}^0,...\} ,...,\{ o_0^t,...,o_{ijk}^t,...\} ,...\}$ satisfies the Markov property. 

\emph{Proof:} To prove \textbf{Lemma 2.}, we need to prove Equation (\ref{eq21}).

\begin{figure}[h] 
	\centering
	\begin{equation}
	\begin{array}{l}	    
	\forall \{ o_0^t,...,o_{ijk}^t,...\} , \\ 
	\displaystyle 	p\{ \{ o_0^t,...,o_{ijk}^t,...\} |\{ o_0^{t - 1},...,o_{ijk}^{t - 1},...\} ,...,\{ o_0^0,...,o_{ijk}^0,...\} \} \\ 
	\displaystyle \qquad \quad = p\{ \{ o_0^t,...,o_{ijk}^t,...\} |\{ o_0^{t - 1},...,o_{ijk}^{t - 1},...\} \}
	\end{array}
	\label{eq21}
	\end{equation}
\end{figure}

According to state space and action space, we know $\{ o_0^t,...,o_{ijk}^t,...\}  - \{ o_0^{t - 1},...,o_{ijk}^{t - 1},...\}  = \{ a_0^{t - 1},...,a_{ijk}^{t - 1},...\}$.

Besides, $\{ a_{ijk}^0,...,a_{ijk}^t\}$ are independent with each other.

So, Equation (\ref{eq22}) are independent with each other.

\begin{figure}[h] 
	\centering
	\begin{equation}
	\begin{array}{l}	    
	\{ \{ o_0^1,...,o_{ijk}^1,...\}  - \{ o_0^0,...,o_{ijk}^0,...\} ,..., \\ 
	\displaystyle 	\{ o_0^t,...,o_{ijk}^t,...\}  - \{ o_0^{t - 1},...,o_{ijk}^{t - 1},...\} ,...\}
	\end{array}
	\label{eq22}
	\end{equation}
\end{figure}

So, $\{ \{ o_0^0,...,o_{ijk}^0,...\} ,...,\{ o_0^t,...,o_{ijk}^t,...\} ,...\}$ has independent incrementality.

Combining with the definition of conditional probability, it can be seen as Equation (\ref{eq23}).

\begin{figure*}[h] 
	\centering
	\begin{equation}	    
	\begin{array}{l}
	\displaystyle p\{ \{ o_0^t,...,o_{ijk}^t,...\} |\{ o_0^{t - 1},...,o_{ijk}^{t - 1},...\} ,...,\{ o_0^0,...,o_{ijk}^0,...\} \} \vspace{3pt} \\ 
	\displaystyle \qquad \quad = \frac{{p\{ \{ o_0^t,...,o_{ijk}^t,...\} ,\{ o_0^{t - 1},...,o_{ijk}^{t - 1},...\} ,...,\{ o_0^0,...,o_{ijk}^0,...\} \} }}{{p\{ \{ o_0^{t - 1},...,o_{ijk}^{t - 1},...\} ,...,\{ o_0^0,...,o_{ijk}^0,...\} \} }} \vspace{3pt} \\
	\displaystyle \qquad \quad = \frac{{p\{ \{ o_0^t,...,o_{ijk}^t,...\}  - \{ o_0^{t - 1},...,o_{ijk}^{t - 1},...\} ,...,\{ o_0^1,...,o_{ijk}^1,...\}  - \{ o_0^0,...,o_{ijk}^0,...\} \} }}{{p\{ \{ o_0^{t - 1},...,o_{ijk}^{t - 1},...\}  - \{ o_0^{t - 2},...,o_{ijk}^{t - 2},...\} ,...,\{ o_0^1,...,o_{ijk}^1,...\}  - \{ o_0^0,...,o_{ijk}^0,...\} \} }} \vspace{3pt} \\ 
	\displaystyle \qquad \quad = \frac{{p\{ \{ o_0^t,...,o_{ijk}^t,...\}  - \{ o_0^{t - 1},...,o_{ijk}^{t - 1},...\} \} ,...,p\{ \{ o_0^1,...,o_{ijk}^1,...\}  - \{ o_0^0,...,o_{ijk}^0,...\} \} }}{{p\{ \{ o_0^{t - 1},...,o_{ijk}^{t - 1},...\}  - \{ o_0^{t - 2},...,o_{ijk}^{t - 2},...\} \} ,...,p\{ \{ o_0^1,...,o_{ijk}^1,...\}  - \{ o_0^0,...,o_{ijk}^0,...\} \} }} \vspace{3pt} \\ 
	\displaystyle \qquad \quad = p\{ \{ o_0^t,...,o_{ijk}^t,...\}  - \{ o_0^{t - 1},...,o_{ijk}^{t - 1},...\} \} 
	\end{array}
	\label{eq23}
	\end{equation}
\end{figure*}

Similarly, we can prove Equation (\ref{eq244}).

\begin{figure}[h] 
	\centering
	\begin{equation}
	\begin{array}{l}	    
	p\{ \{ o_0^t,...,o_{ijk}^t,...\} |\{ o_0^{t - 1},...,o_{ijk}^{t - 1},...\} \}  \\
	\displaystyle \qquad \quad = p\{ \{ o_0^t,...,o_{ijk}^t,...\}  - \{ o_0^{t - 1},...,o_{ijk}^{t - 1},...\} \}
	\end{array}
	\label{eq244}
	\end{equation}
\end{figure}

Next, we can prove Equation (\ref{eq255}).

\begin{figure}[h] 
	\centering
	\begin{equation}
	\begin{array}{l}	    
	p\{ \{ o_0^t,...,o_{ijk}^t,...\} |\{ o_0^{t - 1},...,o_{ijk}^{t - 1},...\} ,...,\{ o_0^0,...,o_{ijk}^0,...\} \}    \\
	\displaystyle \qquad \quad = p\{ \{ o_0^t,...,o_{ijk}^t,...\} |\{ o_0^{t - 1},...,o_{ijk}^{t - 1},...\} \}
	\end{array}
	\label{eq255}
	\end{equation}
\end{figure}

Therefore, $\{ \{ o_0^0,...,o_{ijk}^0,...\} ,...,\{ o_0^t,...,o_{ijk}^t,...\} ,...\}$ satisfies the Markov property.

\subsection{Reward function}

$\left\langle {S,O} \right\rangle  \times A \to r$ represents the expected immediate reward received after the state is transitioned from $\{ {s^{t - 1}},o_0^{t - 1},...,o_{ijk}^{t - 1},...\}$ to $\{ {s^{t}},o_0^{t},...,o_{ijk}^{t},...\}$, due to taking the action $\{ a_0^{t - 1},...,a_{ijk}^{t - 1},...\}$.

The 

In \textbf{Problem 1}, the objective of workers and UAVs is to maximize the number of completed sensing tasks, while the objective of cars is to minimize the urgency of UAVs' power. Therefore, the expected immediate reward ${r^t}$ should include two parts, see Equation (\ref{eq14}). The first part is the sensing tasks completion $taskCp{t^t}$ within the time step $[t,t + 1)$, see Equation (\ref{eq12}). The second part is the reduced urgency $\sum\limits_i {mtigU_i^t}$, which is due to the cars replace the batteries of UAVs at the moment $t$, ${mtigU_i^t}$ see Equation (\ref{eq13}). Equation (\ref{eq13}) is used to calculate the reward for i-th UAV $ua{v_i}$ to alleviate urgency in three situations. (1) When $ua{v_i}$ does not meet any cars, $ua{v_i}$ obtains a reward of 0. (2) When $ua{v_i}$ meets the k-th car $ca{r_k}$ and the remaining battery of $ua{v_i}$ is insufficient to support its flight, $ua{v_i}$ obtains a reward of $1 - \frac{1}{{\mathop e\nolimits^{{\mathop{\rm floor}\nolimits} ({\raise0.7ex\hbox{$1$} \!\mathord{\left/
					{\vphantom {1 {uCs{p_i}}}}\right.\kern-\nulldelimiterspace}
				\!\lower0.7ex\hbox{${uCs{p_i}}$}})} }}$. At this moment, $ua{v_i}$ no longer has flight capability and can only wait for the car, so its power will no longer continue to decrease, its urgency is the highest. (3) When $ua{v_i}$ meets $ca{r_k}$ and the remaining battery of $ua{v_i}$ can still maintain its normal flight, $ua{v_i}$ obtains a reward of $\frac{1}{{{e^{{\rm{floor}}({\raise0.7ex\hbox{${(uPow_i^{t - 1} - uCs{p_i})}$} \!\mathord{\left/
						{\vphantom {{(uPow_i^{t - 1} - uCs{p_i})} {uCs{p_i}}}}\right.\kern-\nulldelimiterspace}
					\!\lower0.7ex\hbox{${uCs{p_i}}$}})}}}} - \frac{1}{{{e^{{\rm{floor}}({\raise0.7ex\hbox{$1$} \!\mathord{\left/
						{\vphantom {1 {uCs{p_i}}}}\right.\kern-\nulldelimiterspace}
					\!\lower0.7ex\hbox{${uCs{p_i}}$}})}}}}$. At this moment, $ua{v_i}$ still has flight capability and can go to a certain place to meet $ca{r_k}$ to replace its battery, so its power will continue to decrease.

\begin{small}
	\begin{equation}
	taskCp{t^t} = \sum\limits_m {task_m^{t-1}}  - \sum\limits_m {task_m^t}
	\label{eq12}
	\end{equation}
\end{small}

\begin{small}
	\begin{equation}
	mtigU_i^t = \left\{ {\begin{array}{*{20}{c}}
		{0,\ {\rm{if }}\ uLoc_i^{t} \ne cLoc_k^{t}} \qquad \qquad \qquad \qquad \qquad \qquad \quad \ \, \vspace{3pt} \\ 
		1 - \frac{1}{{{e^{{\rm{floor}}({\raise0.7ex\hbox{$1$} \!\mathord{\left/
								{\vphantom {1 {uCs{p_i}}}}\right.\kern-\nulldelimiterspace}
							\!\lower0.7ex\hbox{${uCs{p_i}}$}})}}}},\qquad \qquad \qquad \qquad \qquad \qquad \qquad \quad\\
		{\rm{if }}\ uLoc_i^{t} = cLoc_k^{t} \ \& \ uPow_i^{t-1} < uCs{p_i} \qquad \quad \vspace{3pt} \\ 
		\frac{1}{{{e^{{\rm{floor}}({\raise0.7ex\hbox{${(uPow_i^{t - 1} - uCs{p_i})}$} \!\mathord{\left/
								{\vphantom {{(uPow_i^{t - 1} - uCs{p_i})} {uCs{p_i}}}}\right.\kern-\nulldelimiterspace}
							\!\lower0.7ex\hbox{${uCs{p_i}}$}})}}}} - \frac{1}{{{e^{{\rm{floor}}({\raise0.7ex\hbox{$1$} \!\mathord{\left/
								{\vphantom {1 {uCs{p_i}}}}\right.\kern-\nulldelimiterspace}
							\!\lower0.7ex\hbox{${uCs{p_i}}$}})}}}},\\
		{\rm{if }}\ uLoc_i^t = cLoc_k^t \ \& \ uPow_i^{t-1} \ge uCs{p_i} \qquad \, \\
		\end{array}} \right.
	\label{eq13}
	\end{equation}
\end{small}

\begin{small}
	\begin{equation}
	{r^t} = taskCp{t^t} + \sum\limits_i {mtigU_i^t}
	\label{eq14}
	\end{equation}
\end{small}

The traditional expected immediate reward ${r^t}$ see Equation (\ref{eq15}). When any agent chooses an action outside its optional range $agentArriv_{ijk}^t$, ${r^t}$ is set to -10 (negative reward), which punishes current impossible action.

\begin{small}
	\begin{equation}
	{r^t} = \left\{ \begin{array}{l}
	taskCp{t^t} + \sum\limits_i {mtigU_i^t}, \, {\rm{if }}\ \forall a_{ijk}^t \in agentArriv_{ijk}^t \vspace{3pt} \\
	- 10,\ {\rm{else}}
	\end{array} \right.
	\label{eq15}
	\end{equation}
\end{small}

At the moment $t$, when an action is chosen randomly, the probability that we get a positive reward is $pr{o^t} = \prod\limits_i {\frac{{uRge_i^t}}{{AREA}}}  \times \prod\limits_j {\frac{{wRge_j^t}}{{AREA}}}  \times \prod\limits_k {\frac{{cRge_k^t}}{{AREA}}}$. In actual scenarios, $pr{o^t}$ will be extremely small, which leads to the sparse reward. Training models based on the sparse reward is difficult \cite{hare2019dealing}. So, how do we filter out non-feasible actions to avoid negative rewards? During the training process, we use direct logical checks to ensure that each agent's actions are only generated within its optional range $agentArriv_{ijk}^t$. It directly avoids the selection of non-feasible actions, and effectively eliminates the occurrence of negative rewards. Therefore, this paper should use $agentArriv_{ijk}^t$ to filter the non-optional actions and calculate the expected immediate reward based on Equation (\ref{eq14}). It's worth noting that directly adding $taskCp{t^t}$ and $\sum\limits_i {mtigU_i^t}$ in numerical value is not explainable in terms of actual physical meaning. However, in order to estimate the cumulative reward for all types of agents within a single time step, this approach becomes necessary, as outlined in \textbf{Algorithm \ref{algorithm1}} for details. When we consider multiple time steps, it can be simplified into the Equation (\ref{eq12}), as outlined in \textbf{Algorithm \ref{algorithm2}}.

\textbf{Lemma 3.} In \textbf{Problem 1}, UAVs, workers and cars are cooperative.

\emph{Proof:} Workers need to manipulate UAVs to perform the sensing tasks. For the sensing tasks completion $taskCp{t^t}$, UAVs and workers are cooperative.

So, $taskCp{t^t} \propto \{ uLoc_0^t,...,uLoc_i^t,...\}$ and $taskCp{t^t} \propto \{ wLoc_0^t,...,wLoc_j^t,...\}$.

The purpose of cars is to relieve the urgency of the UAVs' power as much as possible. For the reduced urgency $\sum\limits_i {mtigU_i^t}$, $\sum\limits_i {mtigU_i^t}  \propto \{ cLoc_0^t,...,cLoc_k^t,...\}$

Besides, $\{ uPow_0^t,...,uPow_i^t,...\}$ is proportional to the number of working UAVs, e.g., $\{ uPow_0^t,...,uPow_i^t,...\}  \propto \sum\limits_i {mtigU_i^t}$.

And, $\{ uLoc_0^t,...,uLoc_i^t,...\}  \propto \{ uPow_0^t,...,uPow_i^t,...\}$. 

So, $taskCp{t^t} \propto \{ cLoc_0^t,...,cLoc_k^t,...\}$.

Therefore, in \textbf{Problem 1}, UAVs, workers and cars are cooperative.

\section{Methodology}
\label{section5}

In this section, we will introduce the method for collaborative route planning of workers, cars and UAVs for crowdsensing in disaster response. First, we design a heterogeneous multi-agent network framework (MANF) to address the \textbf{Problem 1}. It is worth noting that in MANF, each agent is responsible for controlling either a single UAV, a worker, or a car. Then, we proceed to implement heterogeneous multi-agent route planning algorithms, namely MANF-DNN-RP and MANF-RL-RP, using the MANF framework. MANF-DNN-RP leverages deep learning techniques and incorporates the latest research on UAVs' route planning \cite{liu2019energy, liu2020energy}. On the other hand, MANF-RL-RP is based on MARL and draws inspiration from the QMIX algorithm \cite{rashid2018qmix}.

\subsection{MANF}

\begin{figure}[h]
	\centering
	\begin{minipage}[t]{9cm}
		\setlength{\abovecaptionskip}{0cm}   
		\setlength{\belowcaptionskip}{0cm}   
		\centering 
		\includegraphics[width=9cm]{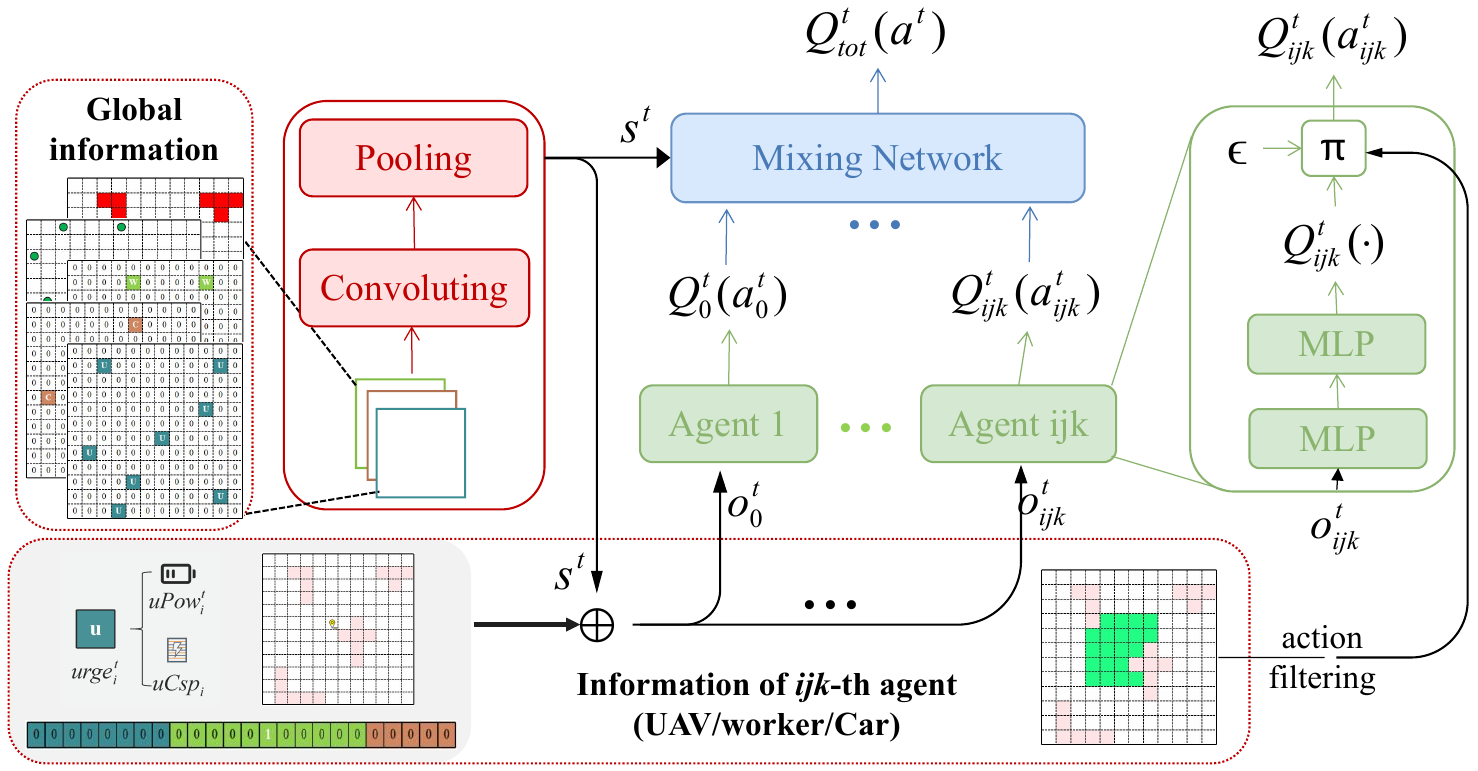}
		\caption{Heterogeneous multi-agent network framework (MANF).}
		\label{figure5}
	\end{minipage}
\end{figure}

Referring to the QMIX algorithm, the MANF consists of two main parts. One is the agent network, which outputs the value $Q_{ijk}^t(a_{ijk}^t)$ for a single agent, while the mixing network takes $Q_{ijk}^t(a_{ijk}^t)$ as input and outputs a joint value $Q_{tot}^t({a^t})$. To maintain consistency between the centralized policy and the decentralized agent policies (e.g., monotonicity), the network parameter weight and offset of the mixing network are calculated through the hypernetworks network \cite{ha2016hypernetworks}. The mixing network weight must be greater than 0, and there is no requirement for mixing network offset. Based on the above description and combined with the research content of this paper, MANF is shown in Figure \ref{figure5}, and there are two points worth noting. (1) We divide the information into two parts, global information and local information of agents. Global information $\{ obstDis{t^t},taskDis{t^t},urgeDis{t^t},workDis{t^t},carDis{t^t}\}$ needs to extract spatial features based on convolutional neural networks and share them with all agents. $\{agentLoc_{ijk}^t,agentI{D_{ijk}},urge_{ijk}^t\}$ in the local information is used to construct the input of the agent network combining with the global state ${s^t}$. $agentArriv_{ijk}^t$ in the local information is used to filter the non-optional actions to avoid negative reward. (2) Due to the decision-making process $\{ \{ o_0^0,...,o_{ijk}^0,...\} ,...,\{ o_0^t,...,o_{ijk}^t,...\} ,...\}$ of a single agent satisfies the Markov property, referring to \textbf{Lemma 2}, we do not need to extract the time series features of agents in the agent network. In addition, since UAVs, workers, and cars are cooperative in \textbf{Problem 1}, referring to \textbf{Lemma 3}, the relationship between the joint actions value $Q_{tot}^t({a^t})$ and the agents' action value $\{ Q_0^t(a_0^t),...,Q_{ijk}^t(a_{ijk}^t),...\}$ satisfies monotonicity. Therefore, in order to ensure the consistency of the joint strategy and the decentralized strategy, the weight of the Mixing Network needs to be non-negative \cite{rashid2018qmix}.

\subsection{MANF-DNN-RP}

Combined with MANF, we implement the heterogeneous multi-agent route planning algorithm MANF-DNN-RP based on deep learning, as shown in \textbf{Algorithm \ref{algorithm1}}. The core idea of the MANF-DNN-RP algorithm is as follows. First, we calculate the expected immediate reward ${r^t}$ for choosing the actions $\{ a_0^t,...,a_{ijk}^t,...\}$ under the current state $\{ o_0^t,...,o_{ijk}^t,...\}$ within the time step $[t,t + 1)$. The expected immediate reward ${r^t}$ consists of two parts, the sensing tasks completion $taskCp{t^t}$ and the reduced urgency $\sum\limits_i {mtigU_i^t}$ within the time step $[t,t + 1)$ , refer to Equation (\ref{eq14}). The whole process is shown in \textbf{Algorithm \ref{algorithm1}}, lines 5-16. Then, we can accurately represent the ternary mapping relationship $< \{ o_0^t,...,o_{ijk}^t,...\}, \{ a_0^t,...,a_{ijk}^t,...\}, {r^t} >$. MANF-DNN-RP can be trained based on the ternary mapping relationship to accurately output the expected immediate reward ${r^t}$ after choosing the actions $\{ a_0^t,...,a_{ijk}^t,...\}$ for the current state $\{ o_0^t,...,o_{ijk}^t,...\}$, as shown in \textbf{Algorithm \ref{algorithm1}}, lines 19-20. Finally, we can compare the expected immediate reward for different actions under the current state $\{ o_0^t,...,o_{ijk}^t,...\}$ based on MANF-DNN-RP, and choose an actions with the largest immediate reward. Repeat the above action selection process until reaching the target moment.

\begin{algorithm}[h]
	\caption{: MANF-DNN-RP} 
	\label{algorithm1}
	\hspace*{0in} \textbf{Input:} 
	$AREA$, $UAV$, $Worker$, $Car$, $TimeLimit$\\
	\hspace*{0in} \textbf{Output:} 
	Spatial Convolutional Network $\rm{cnnSpace}$, Agent Evaluation Network $\rm{evalAgent}$, Mixing Network $\rm{evalMixing}$
	\begin{algorithmic}[1]
		\State Initialize $\rm{cnnSpace}$, $\rm{evalAgent}$, $\rm{evalMixing}$ and experience pool $D$ with size $M$;  
		\While{$\rm{cnnSpace}$, $\rm{evalAgent}$ and $\rm{evalMixing}$ do not converge} 
		\State Index UAVs, workers, and cars with the uniform numbers $\{agentI{D_{ijk}}\}$;
		\For{$t = 0 \to TimeLimit$}  
		\State Get $obstDis{t^t}$ and $taskDis{t^t}$ based on $AREA$;
		\State Get $\{urge_{ijk}^t\}$ and $urgeDis{t^t}$ based on $UAV$; //$urge_{ijk}^t$ of workers and cars are set to 0.
		\State Get $workDis{t^t}$ based on $Worker$;
		\State Get $carDis{t^t}$ based on $Car$;
		\State Get $\{agentLoc_{ijk}^t\}$ and $\{agentArriv_{ijk}^t\}$ based on $UAV$, $Worker$ and $Car$;
		\State ${s^t} = \rm{cnnSpace}(obstDis{t^t},taskDis{t^t},\bullet,\bullet,\bullet)$;
		\Statex \qquad \quad // $\bullet  = urgeDis{t^t}/workDis{t^t}/carDis{t^t}$.
		\State $\{o_{ijk}^t\} = \{\{ {s^t},agentLoc_{ijk}^t,agentI{D_{ijk}},urge_{ijk}^t\}\}$;
		\State $\{ Q_{ijk}^t( \cdot )\}  = \{ {\rm{evalAgent}}(o_{ijk}^t)\}$;
		\State Get $\{ a_{ijk}^t\}$ combining with $\{ Q_{ijk}^t( \cdot )\}$ and $\{agentArriv_{ijk}^t\}$ based on $\varepsilon  - greedy$;
		\State Update $AREA$, $UAV$, $Worker$ and $Car$ based on $\{ a_{ijk}^t\}$;
		\State Get ${r^t}$ based on Equation (\ref{eq14});
		\State $\{s^t,\{o_{ijk}^t\},\{ a_{ijk}^t\},r^t\}$ to $D$;
		\EndFor
		\If{${\rm{length}}(D) \ge M$} 
		\State Randomly sample training data $trainData$ in $D$;
		\Statex \qquad \quad //The following steps are based on $trainData$.
		\State $L(\theta ) = {r^t} - {\rm{evalMixing}}(s^t,\{ Q_{ijk}^t(o_{ijk}^t, a_{ijk}^t)\})$;
		\State Update $\rm{cnnSpace}$, $\rm{evalAgent}$ and $\rm{evalMixing}$ based on $L(\theta )$;
		\EndIf
		\EndWhile
	\end{algorithmic}
\end{algorithm}

\begin{algorithm}[h]
	\caption{: MANF-RL-RP} 
	\label{algorithm2}
	\hspace*{0in} \textbf{Input:} 
	$AREA$, $UAV$, $Worker$, $Car$, $TimeLimit$, $\gamma$\\
	\hspace*{0in} \textbf{Output:} 
	Spatial Convolutional Network $\rm{cnnSpace}$, Agent Evaluation Network $\rm{evalAgent}$, Mixing Network $\rm{evalMixing}$
	\begin{algorithmic}[1]
		\State Initialize $\rm{cnnSpace}$, $\rm{evalAgent}$, $\rm{evalMixing}$, Parameter update frequency $P$ and experience pool $D$ with size $M$;  
		\State Copy $\rm{evalAgent}$ and $\rm{evalMixing}$ to $\rm{tgtAgent}$ and $\rm{tgtMixing}$ respectively;
		\State $CurrStep = 0$;
		\While{$\rm{cnnSpace}$, $\rm{evalAgent}$ and $\rm{evalMixing}$ do not converge} 
		\State Index UAVs, workers, and cars with the uniform numbers $\{agentI{D_{ijk}}\}$;
		\For{$t = 0 \to TimeLimit$} 
		\State Get $s^{t}$, $\{o_{ijk}^{t}\}$ and $\{agentArriv_{ijk}^{t}\}$, referring to lines 7-13 in \textbf{Algorithm \ref{algorithm1}};
		\State $\{ Q_{ijk}^t( \cdot )\}  = \{ {\rm{evalAgent}}(o_{ijk}^t)\}$;
		\State Get $\{ a_{ijk}^t\}$ combining with $\{ Q_{ijk}^t( \cdot )\}$ and $\{agentArriv_{ijk}^t\}$ based on $\varepsilon  - greedy$;
		\State Update $AREA$, $UAV$, $Worker$ and $Car$ based on $\{ a_{ijk}^t\}$;
		\State Get $r^t = {taskCpt^t}$ based on Equation (\ref{eq12});
		\State Get $s^{t+1}$, $\{o_{ijk}^{t+1}\}$ and $\{agentArriv_{ijk}^{t+1}\}$, referring to lines 7-13 in \textbf{Algorithm \ref{algorithm1}};
		\If {$t =  = TimeLimit$}
		\State $te^t = 0$;
		\Else
		\State $te^t = 1$;
		\EndIf
		\State Add $s^t$, $s^{t+1}$, $\{o_{ijk}^t\}$, $\{o_{ijk}^{t+1}\}$, $\{agentArriv_{ijk}^{t+1}\}$, $\{ a_{ijk}^t\}$, $r^t$, $te^t$ to $D$;
		\EndFor
		\If{${\rm{length}}(D) \ge M$} 
		\State Randomly sample training data $trainData$ in $D$;
		\Statex \qquad \quad //The following steps are based on $trainData$.
		\State $\{ mQ_{ijk}^{t + 1}\}  = \{ \max ({\rm{tgtAgent}}(o_{ijk}^{t + 1},\bullet))\}$;
		\Statex \qquad \quad // $\bullet  = agentArriv_{ijk}^{t + 1}$.
		\State ${y^t} = r^t + \gamma \cdot te^t \cdot {\rm{tgtMixing}}({s^{t + 1}},\{ mQ_{ijk}^{t + 1}\} )$;
		\State $L(\theta ) = {y^t} - {\rm{evalMixing}}(s^t,\{ Q_{ijk}^t(o_{ijk}^t, a_{ijk}^t)\})$;
		\State Update $\rm{cnnSpace}$, $\rm{evalAgent}$ and $\rm{evalMixing}$ based on $L(\theta )$;
		\State $CurrStep = CurrStep + 1$
		\EndIf
		\If{$CurrStep\ \%\ P =  = 0$}
		\State Copy $\rm{evalAgent}$ and $\rm{evalMixing}$ to $\rm{tgtAgent}$ and $\rm{tgtMixing}$ respectively;
		\EndIf
		\EndWhile
	\end{algorithmic}
\end{algorithm}

\subsection{MANF-RL-RP}

The MANF-DNN-RP algorithm only focuses on how to obtain the optimal action $\{ a_0^t,...,a_{ijk}^t,...\}$ under the current state $\{ o_0^t,...,o_{ijk}^t,...\}$ within the time step $[t,t + 1)$, while ignores the long-term impact of $\{ o_0^t,...,o_{ijk}^t,...\}$ on subsequent action selection. Therefore, we implement a heterogeneous multi-agent route planning algorithm MANF-RL-RP based on MARL, which can take into account the long-term impact of $\{ o_0^t,...,o_{ijk}^t,...\}$. Based on Equation (\ref{eq14}), we can estimate the total expected immediate reward ${G^t}$ within the time step $[t,TimeLimit]$. ${G^t}$ represents the cumulative sum of immediate reward during the time step $[t,TimeLimit]$, referring to Equation (\ref{eq16}). $\gamma$ is the discount factor, where a higher value indicates a greater emphasis on future immediate reward, while a lower value indicates a greater emphasis on immediate reward in the near term.

\begin{small}
	\begin{equation}
	\begin{array}{l}
	{G^t} = {r^t} + \gamma {r^{t + 1}} + ... + {\gamma ^{TimeLimit - t}}{r^{TimeLimit}}\\
	\quad \ \, = (taskCp{t^t} + \sum\limits_i {mtigU_i^t} ) + ... +\\
	{\gamma ^{TimeLimit - t}}(taskCp{t^{TimeLimit}} + \sum\limits_i {mtigU_i^{TimeLimit}} )
	\end{array}
	\label{eq16}
	\end{equation}
\end{small}

However, the impact of the cars on the task completion rate is delayed, referring to \textbf{Lemma 3}. Replacing the UAVs' batteries with the cars at the current moment can reduce the occurrence of the UAVs halting operations in future moments due to insufficient power. In other words, $\sum\limits_i {mtigU_i^t}$ will be reflected on $\{ taskCp{t^{t + 1}},...,taskCp{t^{TimeLimit}}\}$. In addition, the optimization objective of \textbf{Problem 1} is to maximize the task completion $\sum\limits_m {task_m^0}  - \sum\limits_m {task_m^{TimeLimit}}$, which is inconsistent with the total expected immediate reward ${G^t}$ in Equation (\ref{eq16}). Therefore, if we compute the immediate reward based on Equation (\ref{eq14}), $\{ \sum\limits_i {mtigU_i^t} ,...,\sum\limits_i {mtigU_i^{TimeLimit}} \}$ may affect the ability of agents to make optimal decisions in MANF-RL-RP algorithm. Therefore, we can can calculate immediate reward based on Equation (\ref{eq12}), and further simplify the total expected immediate reward ${G^t}$ to $G_{task}^t$, see Equation (\ref{eq17}), which is fitter the optimization objective in \textbf{Problem 1}. Finally, we can refer to the optimization process of standard reinforcement learning algorithms for the MANF-RL-RP algorithm, as shown in \textbf{Algorithm \ref{algorithm2}}. Additionally, it is worth noting that setting up a target network and an evaluation network can address the issue of training instability in MANF-RL-RP. During the training process of MANF-RL-RP, using a single neural network can lead to two problems: Firstly, the target values are estimated based on the single neural network, and these estimated values may have biases. Secondly, updating the network leads to modifications in the estimated values, thereby exacerbating the disparity between the target values and the estimated values. To address these problems, MANF-RL-RP incorporates a target network and an evaluation network. The target network is periodically updated based on the evaluation network to slow down the rate of target value changes. In contrast, MANF-DNN-RP utilizes target values that are derived from objective real-world environments, ensuring stability and freedom from bias.

\begin{small}
	\begin{equation}
	\begin{array}{l}
	G_{task}^t = taskCp{t^t} + ... + {\gamma ^{TimeLimit - t}}taskCp{t^{TimeLimit}}
	\end{array}
	\label{eq17}
	\end{equation}
\end{small}

\begin{table*}[hb]
	\centering
	\footnotesize
	\caption{Simulated data}
	\label{tab1}
	\setlength{\tabcolsep}{1.4mm}{
		\begin{tabular}{c|p{1.2cm}<{\centering}p{2.7cm}<{\centering}p{2.6cm}<{\centering}p{2.2cm}<{\centering}p{1.5cm}<{\centering}p{1.7cm}<{\centering}p{1.5cm}<{\centering}}
			\toprule
			data ID & initial locations & agent number & area number & tasks number (distribution) & obstacle number & movable radius & powers consumed\\
			\midrule
			(1) & same & 10/25/5 & 16*16 & 120(random) & 20 & 8/3/5 & 0.3\\
			\midrule
			(2) & random & 10/25/5 & 16*16 & 120(random) & 20 & [7,9]/[2,4]/[4,6] & [0.2,0.4]\\
			\midrule
			(3) & check-in & 10/25/5 & 16*16 & 120(random) & 20 & [7,9]/[2,4]/[4,6] & [0.2,0.4]\\
			\midrule
			(4) & check-in & 10/25/5 & 16*16 & 120(check-in) & 20 & [7,9]/[2,4]/[4,6] & [0.2,0.4]\\
			\midrule
			(5) & random & (8/20/4),(10/25/5),(12/30/6) & 16*16 & 120(random) & 20 & [7,9]/[2,4]/[4,6] & [0.2,0.4]\\
			\midrule
			(6) & random & 10/25/5 & (12*12),(16*16),(20*20) & 120(random) & 20 & [7,9]/[2,4]/[4,6] & [0.2,0.4]\\
			\midrule
			(7) & random & 10/25/5 & 16*16 & 100/120/140(random) & 20 & [7,9]/[2,4]/[4,6] & [0.2,0.4]\\
			\midrule
			(8) & random & 10/25/5 & 16*16 & 120(random) & 20/40/60 & [7,9]/[2,4]/[4,6] & [0.2,0.4]\\
			\bottomrule
	\end{tabular}}
\end{table*}

\begin{table*}[hb]
	\centering
	\footnotesize
	\caption{Experimental group settings}
	\label{tab2}
	\setlength{\tabcolsep}{1.6mm}{
		\begin{tabular}{c|ccc}
			\toprule
			group ID & data ID & $TimeLimit$ & algorithm (abbreviation)\\
			\midrule
			(1) & (1) / (2) / (3) / (4) & 9 & MANF-DNN-RP-temp / MANF-DNN-RP / MANF-RL-RP-temp / MANF-RL-RP\\
			\midrule
			(2) & (1) / (2) / (3) / (4) & 6 / 9 / 12 & Greedy-SC-RP / MANF-DNN-RP / MANF-RL-RP\\
			\midrule
			(3) & (5) / (6) / (7) / (8) & 9 & Greedy-SC-RP / MANF-DNN-RP / MANF-RL-RP\\
			\bottomrule
	\end{tabular}}
\end{table*}

\section{EVALUATION}
\label{section6}

\subsection{Data set}

We conduct experimental evaluations based on simulated data, which includes discrete area set $AREA$, UAVs set $UAV$, workers set $Worker$, and cars set $Car$. To conduct experimental evaluations more objectively and comprehensively, the following points are worth noting in the simulated data, refer to \cite{liu2020multi, liu2019distributed, liu2019energy, 9641503}.

\begin{figure}[h]
	\vspace{-0.3cm} 
	\setlength{\abovecaptionskip}{0.2cm}   
	\setlength{\belowcaptionskip}{-0.3cm}   
	\subfigcapskip=-0.1cm 
	\centering
	\subfigure[random distribution]{
		\begin{minipage}[t]{0.5\linewidth}
			\centering
			\includegraphics[width=4.3cm]{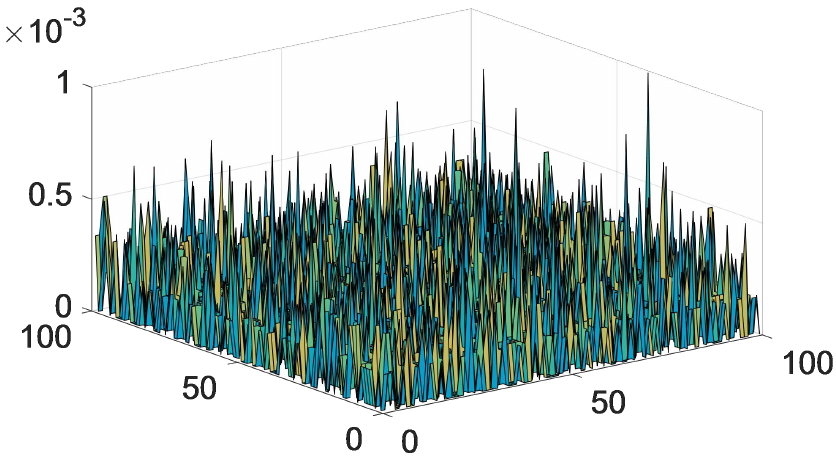}
			\label{figure51}
		\end{minipage}%
	}%
	\subfigure[check-in empirical distribution]{
		\begin{minipage}[t]{0.5\linewidth}
			\centering
			\includegraphics[width=4.3cm]{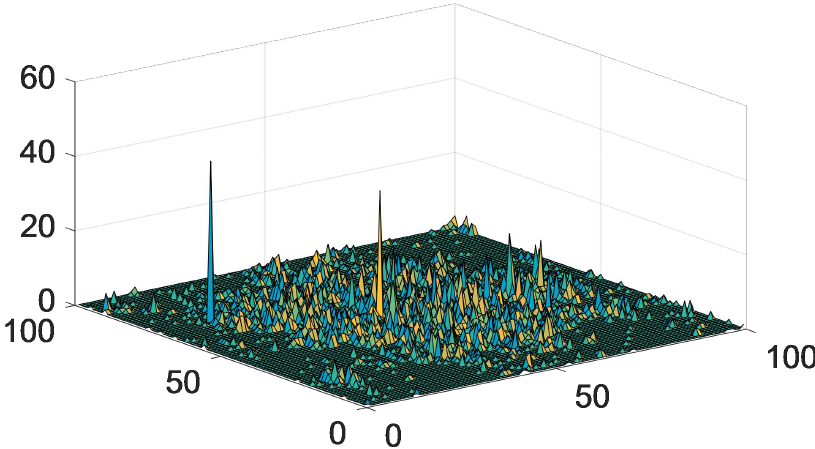}
			\label{figure52}
		\end{minipage}%
	}%
	\centering
	\caption{Geographical distributions of agents and sensing tasks.}
	\label{figure512}
\end{figure}

(1) There cannot be both obstacles and sensing tasks in an area. The locations of the obstacles satisfy a random distribution. The locations of the sensing tasks satisfy random distribution or check-in empirical distribution , as shown in Figure \ref{figure512}. Note: The check-in data records the human position in the real-world environment. The check-in empirical distribution is simulated based on check-in data \cite{2011Friendship}, which represents a density distribution of human in the geographic locations.

(2) The initial locations of all agents (e.g., UAVs, workers and cars) satisfy the following three distributions, as indicated in reference \cite{han2021online}. (1) The initial locations of all agents are same; (2) The initial locations of all agents satisfy random distribution, as shown in Figure \ref{figure51}; (3) The initial locations of all agents satisfy the check-in empirical distribution, as shown in Figure \ref{figure52}. 

(3) For agents of the same type, the permissible range of movement can be generated in two ways: (1) all agents have the same fixed movable radius; (2) the movable radius of each agent is randomly generated within a specified interval. It should be noted that the permissible ranges of movement for agents are determined based on the movable radius and environmental information, excluding areas with obstacles.

(4) The powers consumed by UAVs in a time step are generated in two ways. (1) The powers of UAVs are same; (2) The powers of UAVs are randomly set. Note that the initial powers of all UAVs are 1.

Based on the above requirements, we simulated 8 sets of data, as shown in Table \ref{tab1}. In addition, based on the simulated data, we set up 3 groups of experiments, as shown in Table \ref{tab2}.

\subsection{Experiment Setup}

Since neural networks are not the focus of our research, we use the same neural network in the control experiments. Study \cite{1989Approximation} has shown that a layer of the neural network can fit any function. Therefore, all methods in this paper use neural networks with one hidden layer, and the hidden layer nodes are 10 times as large as the input layer nodes. In addition, the spatial convolutional network $\rm{cnnSpace}$ only contains one convolutional layer and one average pooling layer. All activation functions are relu() in this paper. The other experimental hyperparameters are shown in Table \ref{tab3}. It is worth noting that the earlier the sensing tasks are completed, the better. Therefore, according to experience, we set $\gamma$ to a smaller value to pay attention to the sensing tasks completion at the current moment as much as possible. We are confident that we can further improve the performance of the algorithm by tuning parameters, but it is not helpful to study the core issues in this paper.

\begin{table}[h]
	\centering
	\footnotesize
	\caption{Experimental hyperparameters}
	\label{tab3}
	\setlength{\tabcolsep}{1.6mm}{
		\begin{tabular}{ccccc}
			\toprule
			learning rate & $\gamma$ & $P$ & $M$ & $\varepsilon$ \\
			0.0001 & 0.7 & 200 & 5000 & 1/0.1/32\\
			\midrule
			batch size & optimizer & / & in\_channels & out\_channels \\
			32 & RMSprop & / & 5 & 10 \\
			\midrule
			kernel\_size & stride & padding & dilation & Pool\_size\\
			3 & 1 & 1 & 1 & 2 \\
			\bottomrule
	\end{tabular}}
\end{table}

\subsection{Baselines and evaluation indicator}

Baselines are as follows.

(1) We evaluated the most relevant work \cite{wang2022task, li2019three}. Given that our research problem differs from existing works, we have made modifications to these approaches in order to address our specific problem, resulting in the development of the Greedy-SC-RP algorithm. The Greedy-SC-RP algorithm employs a greedy approach to sequentially plan routes for UAVs, workers, and cars. The implementation process of the algorithm can be outlined in the following three steps.
	
First, calculate the sum of Euclidean distances $Dis_{ii}^t$ between different locations $locOpt_{ii}^t$ within the optional range $agentArriv_{i}^t$ of the UAV $uav_i^t$ and all locations of sensing tasks $tas{k_m},(m = 1,2,...)$, as shown in Equation (\ref{eq24}). The function "ED()" denotes the calculation of the Euclidean distance between two locations. Select the location with the minimum Euclidean distance $Dis_{ii}^t$ as the location for the UAV $uav_i^{t+1}$ within the time step $[t,t+1]$. Repeat the above process until the next location for all UAVs are computed.

\begin{small}
	\begin{equation}
	\begin{array}{l}
	Dis_{ii}^t = \sum\limits_m {{\mathop{\rm ED}\nolimits} (locOpt_{ii}^t,tas{k_m})} ,locOpt_{ii}^t \in agentArriv_i^t
	\end{array}
	\label{eq24}
	\end{equation}
\end{small}

Then, calculate the sum of Euclidean distances $Dis_{jj}^t$ between different locations $locOpt_{jj}^t$ within the optional range $agentArriv_{j}^t$ of the worker $wkr_j^t$ and all locations of sensing tasks $tas{k_m},(m = 1,2,...)$, see Equation (\ref{eq25}). Select the location with the minimum Euclidean distance $Dis_{jj}^t$ as the location for the worker $wkr_j^{t+1}$ within the time step $[t,t+1]$. Repeat the above process until the next location for all workers are computed.

\begin{small}
	\begin{equation}
	\begin{array}{l}
	Dis_{jj}^t = \sum\limits_m {{\mathop{\rm ED}\nolimits} (locOpt_{jj}^t,tas{k_m})} ,locOpt_{jj}^t \in agentArriv_j^t
	\end{array}
	\label{eq25}
	\end{equation}
\end{small}

Finally, calculate the sum of Euclidean distances $Dis_{kk}^t$ between different locations $locOpt_{kk}^t$ within the optional range $agentArriv_{k}^t$ of the car $car_k^t$ and locations of UAVs $uav_i^{t+1},(i = 1,2,...)$ at moment $t+1$ , see Equation (\ref{eq26}). Select the location with the minimum Euclidean distance $Dis_{kk}^t$ as the location for the car $car_k^{t+1}$ within the time step $[t,t+1]$. Repeat the above process until the next location for all cars are computed.

\begin{small}
	\begin{equation}
	\begin{array}{l}
	Dis_{kk}^t = \sum\limits_i {{\mathop{\rm ED}\nolimits} (locOpt_{kk}^t,uav_i^{t + 1})} , locOpt_{kk}^t \in agentArriv_k^t
	\end{array}
	\label{eq26}
	\end{equation}
\end{small}

(2) MANF-DNN-RP-temp: Remove the spatial convolutional network $\rm{cnnSpace}$ in the MANF-DNN-RP algorithm.

(3) MANF-RL-RP-temp: Change the expected immediate reward from $taskCp{t^t}$ to $taskCp{t^t} + \sum\limits_i {mtigU_i^t}$ in the MANF-RL-RP algorithm.

According to \textbf{Problem 1}, we use the task completion rate within the time step $[0,TimeLinit]$ as evaluation indicator, see Equation (\ref{eq18}).

\begin{small}
	\begin{equation}
	\begin{array}{l}
	taskCptRate = \frac{{\sum\limits_m {task_m^0}  - \sum\limits_m {task_m^{TimeLimit}} }}{{\sum\limits_m {task_m^0} }}
	\end{array}
	\label{eq18}
	\end{equation}
\end{small}

\subsection{Experiment Results}

\subsubsection{Verifying the improvement of methods}

The experimental setting refers to group (1) in Table \ref{tab2}. The experimental results are shown in Table \ref{tab4}. Compared to MANF-DNN-RP-temp, the utilization of MANF-DNN-RP, which expands the spatial convolutional network $\rm{cnnSpace}$ to extract spatial features of global information, leads to a substantial enhancement in the average task completion rate by 2.50\%. Introducing spatial convolutional neural networks effectively extracts and preserves the two-dimensional spatial distribution features of the original global variables. On the other hand, algorithms that do not utilize spatial convolutional networks forcibly transform the two-dimensional distribution of global variables into one-dimensional vectors, making it difficult to capture and retain the inherent two-dimensional distribution characteristics of the global variables. When comparing MANF-RL-RP-temp to MANF-RL-RP, the simplification of the expected immediate reward from $taskCp{t^t} + \sum\limits_i {mtigU_i^t}$ to $taskCp{t^t}$ brings about a significant increase in the average task completion rate by 12.71\%. The impact of the cars on the task completion rate is delayed, referring to \textbf{Lemma 3}. Replacing the UAVs' batteries with the cars at the current moment can reduce the occurrence of the UAVs halting operations in future moments due to insufficient power. In other words, $\sum\limits_i {mtigU_i^t}$ will be reflected on $\{ taskCp{t^{t + 1}},...,taskCp{t^{TimeLimit}}\}$. In addition, the optimization objective of \textbf{Problem 1} is to maximize the task completion $\sum\limits_m {task_m^0}  - \sum\limits_m {task_m^{TimeLimit}}$, which is inconsistent with the total expected immediate reward ${G^t}$ in Equation (\ref{eq16}). Therefore, if we compute the immediate reward based on $taskCp{t^t} + \sum\limits_i {mtigU_i^t}$, $\{ \sum\limits_i {mtigU_i^t} ,...,\sum\limits_i {mtigU_i^{TimeLimit}} \}$ may affect the ability of agents to make optimal decisions in MANF-RL-RP algorithm. Therefore, we can can calculate immediate reward based on $taskCp{t^t}$.

\begin{table}[h]
	\centering
	\footnotesize
	\caption{Performance comparison based on different data}
	\label{tab4}
	\setlength{\tabcolsep}{1.5mm}{
		\begin{tabular}{c|p{1.9cm}<{\centering}p{1.5cm}<{\centering}p{1.8cm}<{\centering}p{1.5cm}<{\centering}}
			\toprule
			dataID & MANF-DNN-RP-temp & MANF-DNN-RP & MANF-RL-RP-temp & MANF-RL-RP \\
			\midrule
			(1) & 0.4667 & 0.4917 & 0.4500 & 0.6417 \\
			\midrule
			(2) & 0.5083 & 0.5417 & 0.4250 & 0.5833 \\
			\midrule
			(3) & 0.5833 & 0.6083 & 0.5417 & 0.6250 \\
			\midrule
			(4) & 0.5583 & 0.5750 & 0.5917 & 0.6667 \\
			\bottomrule
	\end{tabular}}
\end{table}

\subsubsection{Changing $TimeLimit$ under the different data distribution}

\begin{figure}[h]
	\vspace{-0.3cm} 
	\setlength{\abovecaptionskip}{0.2cm}   
	\setlength{\belowcaptionskip}{-0.3cm}   
	\subfigcapskip=-0.1cm 
	\centering
	\subfigure[changing \emph{TimeLimit}, data (1)]{
		\begin{minipage}[t]{0.5\linewidth}
			\centering
			\includegraphics[width=4.3cm]{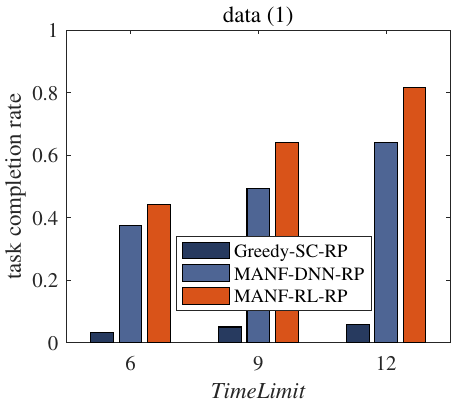}
			\label{figure61}
		\end{minipage}%
	}%
	\subfigure[changing \emph{TimeLimit}, data (2)]{
		\begin{minipage}[t]{0.5\linewidth}
			\centering
			\includegraphics[width=4.3cm]{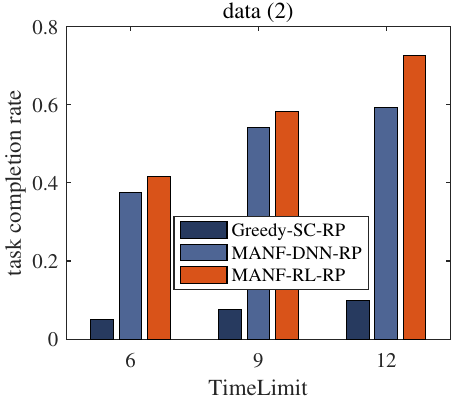}
			\label{figure62}
		\end{minipage}%
	}%
	
	\subfigure[changing \emph{TimeLimit}, data (3)]{
		\begin{minipage}[t]{0.5\linewidth}
			\centering
			\includegraphics[width=4.3cm]{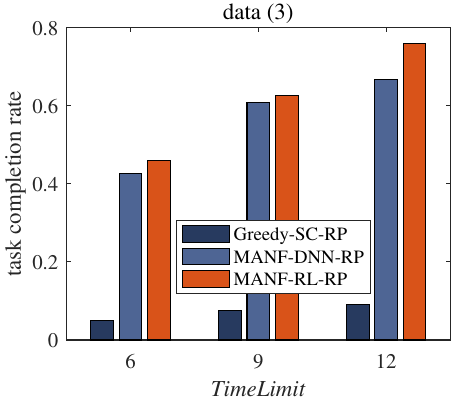}
			\label{figure63}
		\end{minipage}%
	}%
	\subfigure[changing \emph{TimeLimit}, data (4)]{
		\begin{minipage}[t]{0.5\linewidth}
			\centering
			\includegraphics[width=4.3cm]{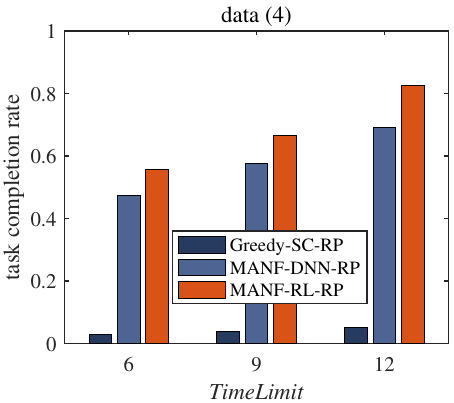}
			\label{figure64}
		\end{minipage}%
	}%
	\centering
	\caption{Changing $TimeLimit$ under the different data distribution.}
	\label{figure6}
\end{figure}

The experimental setting refers to group (2) in Table \ref{tab2}. The experimental results are shown in Figure \ref{figure6}. With the increase of upper limit of sensing time $TimeLimit$, the task completion rate of all methods is gradually increasing. The increase in $TimeLimit$ indicates that UAVs, workers, and cars can spend more time executing the sensing tasks, and more time results in higher task completion rate. In addition, under different $TimeLimit$, the task completion rate of MANF-RL-RP is significantly higher than that of MANF-DNN-RP and Greedy-SC-RP. When $TimeLimit$ is 6, the task completion rate is increased by 5.63\% and 42.86\% on average, respectively. When $TimeLimit$ is 9, the task completion rate is increased by 7.50\% and 56.94\% on average, respectively. When $TimeLimit$ is 12, the task completion rate is increased by 13.33\% and 70.60\% on average, respectively. Since Greedy-SC-RP greedily plans routes for agents in turn (i.e., UAVs, workers and cars), it is difficult to capture the collaboration between agents, and does not perform well. Next, MANF-DNN-RP is essentially a greedy idea, which gives priority to the best cooperation of all agents at the current moment, regardless of the long-term impact of the current choice on subsequent decisions. MANF-RL-RP is implemented based on reinforcement learning, which can take into account the long-term impact of current choices on subsequent decisions. For a more detailed analysis, please refer to Section \ref{SectionAD}.

\subsubsection{Changing the number of agents, areas, sensing tasks or obstacles}

\begin{figure}[h]
	\vspace{-0.3cm} 
	\setlength{\abovecaptionskip}{0.2cm}   
	\setlength{\belowcaptionskip}{-0.3cm}   
	\subfigcapskip=-0.1cm 
	\centering
	\subfigure[Changing the number of agents.]{
		\begin{minipage}[t]{0.5\linewidth}
			\centering
			\includegraphics[width=4.3cm]{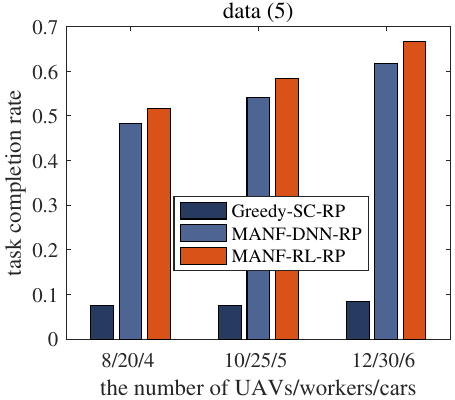}
			\label{figure71}
		\end{minipage}%
	}%
	\subfigure[Changing the number of areas]{
		\begin{minipage}[t]{0.5\linewidth}
			\centering
			\includegraphics[width=4.3cm]{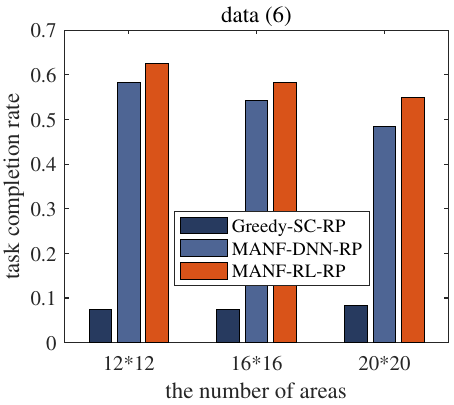}
			\label{figure72}
		\end{minipage}%
	}%
	
	\subfigure[Changing the number of sensing tasks]{
		\begin{minipage}[t]{0.5\linewidth}
			\centering
			\includegraphics[width=4.3cm]{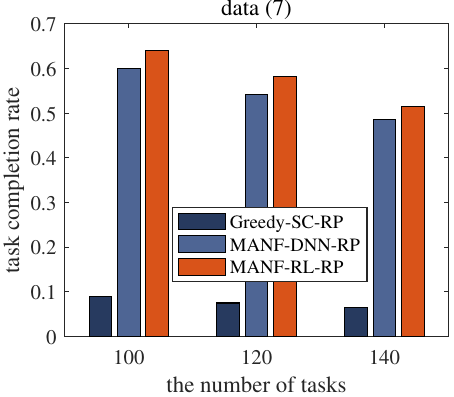}
			\label{figure73}
		\end{minipage}%
	}%
	\subfigure[Changing the number of obstacles]{
		\begin{minipage}[t]{0.5\linewidth}
			\centering
			\includegraphics[width=4.3cm]{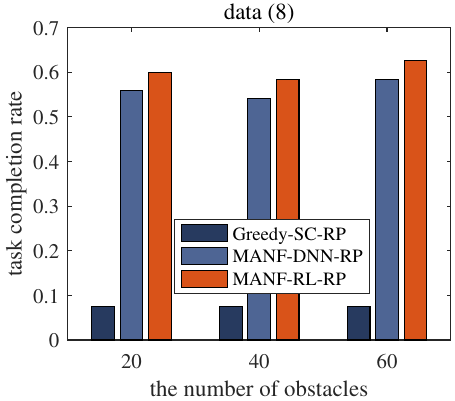}
			\label{figure74}
		\end{minipage}%
	}%
	\centering
	\caption{Changing the number of agents, areas, sensing tasks or obstacles}
	\label{figure7}
\end{figure}

The experimental setting refers to group (3) in Table \ref{tab2}. The experimental results are shown in Figure \ref{figure7}. With the increase of agents (i.e., UAVs, workers and cars), the task completion rate of all methods is gradually increasing, as shown in Figure \ref{figure71}. The increase of agents indicates that there are more UAVs, workers and cars to perform sensing tasks simultaneously, which can complete more sensing tasks per unit time, finally leading to a higher task completion rate. With the increase of the number of areas, the task completion rate of all methods is gradually decreasing, as shown in Figure \ref{figure72}. The increase of areas leads to a more sparse distribution of sensing tasks, and the mobility of agents is limited, which inevitably leads to lower task completion rates. With the increase of the number of sensing tasks, the task completion rate of all methods is gradually decreasing, as shown in Figure \ref{figure73}. There is an upper limit to the number of sensing tasks that a given number of agents can complete within a limited time. With the increase of the number of obstacles, there is a slight increase in the task completion rate of all methods, as shown in Figure \ref{figure74}. The increase of obstacles reduces the sparsity of sensing task distribution. The high density distribution of sensing tasks is conducive to increasing the task completion rate.

\subsection{Analysis and Discussion}
\label{SectionAD}

Since Greedy-SC-RP greedily plans routes for agents in turn (i.e., UAVs, workers and cars), it is difficult to capture the collaboration between agents, and does not perform well. Compared with MANF-DNN-RP-temp, MANF-DNN-RP expands the spatial convolutional network $\rm{cnnSpace}$ to extract spatial features of global information, and performs batter \cite{liu2020energy, liu2019distributed, liu2019energy, wang2021energy, liu2020curiosity}. Next, we will compare MANF-RL-RP, MANF-DNN-RP and MANF-RL-RP-temp to illustrate the advantages of MANF-RL-RP.

\subsubsection{MANF-RL-RP VS. MANF-DNN-RP}

To explain the difference between MANF-RL-RP and MANF-DNN-RP more clearly, we need to calculate the task completion rate per unit time $taskCptRatePer^t$, see Equation (\ref{eq19}).

\begin{small}
	\begin{equation}
	\begin{array}{l}
	taskCptRatePer^t = \sum\limits_m {task_m^t}  - \sum\limits_m {task_m^{t+1}}, {t \in [1,TimeLimit]}
	\end{array}
	\label{eq19}
	\end{equation}
\end{small}

\begin{figure}[h]
	\centering
	\begin{minipage}[t]{9cm}
		\setlength{\abovecaptionskip}{0cm}   
		\setlength{\belowcaptionskip}{0cm}   
		\centering 
		\includegraphics[width=8.5cm]{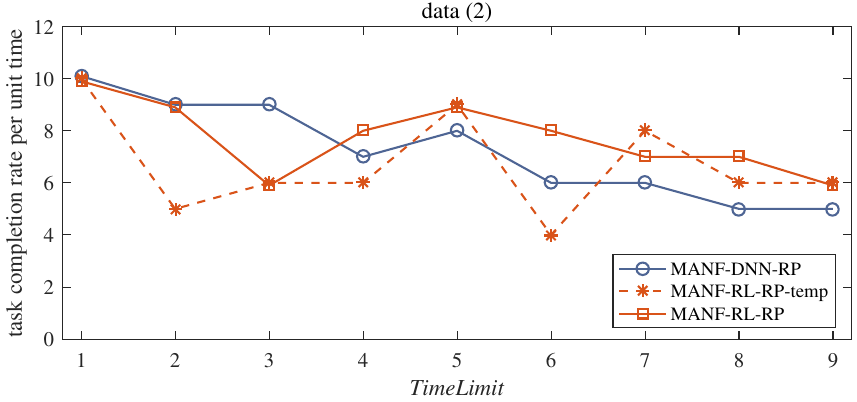}
		\caption{$taskCp{t^t}$ at different moments, $TimeLimit = 9$.}
		\label{figure8}
	\end{minipage}
\end{figure}

MANF-RL-RP performs better than MANF-DNN-RP mainly for two reasons. (1) MANF-DNN-RP is essentially a greedy idea, which gives priority to the best cooperation of all agents at the current moment, regardless of the long-term impact of the current choice on subsequent decisions. MANF-RL-RP is implemented based on reinforcement learning, which can take into account the long-term impact of current choices on subsequent decisions. As shown in Figure \ref{figure8}, the task completion rate per unit time $taskCptRatePer^t$ of MANF-DNN-RP is slightly higher than $taskCptRatePer^t$ of MANF-RL-RP at $t \in [1,3]$, and then significantly worse than $taskCptRatePer^t$ of MANF-RL-RP at $t \in [4,9]$. (2) However, MANF-RL-RP-temp is also implemented based on reinforcement learning. As shown in Figure \ref{figure8}, Why is the task completion rate per unit time $taskCptRatePer^t$ of MANF-RL-RP-temp almost always worse than $taskCptRatePer^t$ of MANF-RL-RP? To take into account the promotion of cars on carrying out sensing tasks, MANF-DNN-RP uses $taskCp{t^t} + \sum\limits_i {mtigU_i^t}$ as the greed indicator, which is not perfect fit with the optimization goal of \textbf{Problem 1}. MANF-RL-RP uses $taskCp{t^t}$ to estimate the expected immediate reward, which can well fit the optimization goal of \textbf{Problem 1}. For detailed explanation, please refer to Section \ref{section6E}.

\subsubsection{MANF-RL-RP VS. MANF-RL-RP-temp}
\label{section6E}

Table \ref{tab5} records the remaining power of UAVs at different moments, in which \colorbox{cyan}{\color{black}cyan mark} indicates that the batteries of the UAVs have been replaced. Based on Table \ref{tab5}, we can get two differences between MANF-RL-RP-temp and MANF-RL-RP. (1) When the batteries of UAVs are replaced, the power of UAVs in MANF-RL-RP-temp is usually higher than that in MANF-RL-RP. (2) The battery replacement frequency of UAVs in MANF-RL-RP-temp (i.e., 42 times) is higher than that of UAVs in MANF-RL-RP (i.e., 30 times). We should replace the batteries of UAVs without affecting performing the sensing tasks, rather than replacing their batteries when the power of UAVs is still high. In addition, frequently meeting cars to replace the batteries of UAVs may seriously affect the efficiency of UAVs in performing sensing tasks. Therefore, it is better to choose $taskCp{t^t}$ (i.e., MANF-RL-RP) to calculate the expected immediate reward than $taskCp{t^t} + \sum\limits_i {mtigU_i^t}$ (i.e., MANF-RL-RP-temp) in \textbf{Problem 1}.

\begin{table*}[h]
	\centering
	\footnotesize
	\caption{Power of UAVs at different moments, $TimeLimit = 9$.}
	\label{tab5}
	\setlength{\tabcolsep}{1.5mm}{
		\begin{tabular}{|c|p{0.55cm}<{\centering}p{0.55cm}<{\centering}p{0.55cm}<{\centering}p{0.55cm}<{\centering}p{0.55cm}<{\centering}p{0.55cm}<{\centering}p{0.55cm}<{\centering}p{0.55cm}<{\centering}p{0.55cm}<{\centering}|p{0.55cm}<{\centering}p{0.55cm}<{\centering}p{0.55cm}<{\centering}p{0.55cm}<{\centering}p{0.55cm}<{\centering}p{0.55cm}<{\centering}p{0.55cm}<{\centering}p{0.55cm}<{\centering}p{0.55cm}<{\centering}|}
			\toprule
			method (data (2)) & \multicolumn{9}{c|}{MANF-RL-RP-temp} & \multicolumn{9}{c|}{MANF-RL-RP} \\ \hline
			\diagbox{UAV's ID}{moment $t$} & 1 & 2 & 3 & 4 & 5 & 6 & 7 & 8 & 9 & 1 & 2 & 3 & 4 & 5 & 6 & 7 & 8 & 9 \\
			\midrule
			1 & 0.71 & \colorbox{cyan}{\color{black}1} & \colorbox{cyan}{\color{black}1} & 0.71 & 0.42 & \colorbox{cyan}{\color{black}1} & 0.71 & 0.42 & \colorbox{cyan}{\color{black}1} & 0.71 & \colorbox{cyan}{\color{black}1} & 0.71 & \colorbox{cyan}{\color{black}1} & \colorbox{cyan}{\color{black}1} & \colorbox{cyan}{\color{black}1} & 0.71 & \colorbox{cyan}{\color{black}1} & 0.71 \\
			2 & 0.64 & \colorbox{cyan}{\color{black}1} & 0.64 & \colorbox{cyan}{\color{black}1} & 0.64 & \colorbox{cyan}{\color{black}1} & \colorbox{cyan}{\color{black}1} & 0.64 & \colorbox{cyan}{\color{black}1} & 0.64 & 0.28 & 0.28 & 0.28 & 0.28 & \colorbox{cyan}{\color{black}1} & 0.64 & \colorbox{cyan}{\color{black}1} & 0.64 \\
			3 & 0.65 & \colorbox{cyan}{\color{black}1} & 0.65 & \colorbox{cyan}{\color{black}1} & 0.65 & \colorbox{cyan}{\color{black}1} & 0.65 & \colorbox{cyan}{\color{black}1} & 0.65 & 0.65 & 0.30 & \colorbox{cyan}{\color{black}1} & 0.65 & 0.30 & 0.30 & \colorbox{cyan}{\color{black}1} & 0.65 & 0.30 \\
			4 & 0.67 & \colorbox{cyan}{\color{black}1} & 0.67 & \colorbox{cyan}{\color{black}1} & 0.67 & \colorbox{cyan}{\color{black}1} & \colorbox{cyan}{\color{black}1} & \colorbox{cyan}{\color{black}1} & 0.67 & 0.67 & 0.34 & \colorbox{cyan}{\color{black}1} & 0.67 & 0.34 & \colorbox{cyan}{\color{black}1} & \colorbox{cyan}{\color{black}1} & 0.67 & \colorbox{cyan}{\color{black}1} \\
			5 & 0.61 & \colorbox{cyan}{\color{black}1} & 0.61 & \colorbox{cyan}{\color{black}1} & 0.61 & \colorbox{cyan}{\color{black}1} & 0.61 & \colorbox{cyan}{\color{black}1} & 0.61 & 0.61 & 0.22 & 0.22 & \colorbox{cyan}{\color{black}1} & 0.61 & 0.22 & 0.22 & \colorbox{cyan}{\color{black}1} & 0.61 \\
			6 & 0.74 & \colorbox{cyan}{\color{black}1} & 0.74 & 0.48 & \colorbox{cyan}{\color{black}1} & \colorbox{cyan}{\color{black}1} & 0.74 & 0.48 & \colorbox{cyan}{\color{black}1} & 0.74 & 0.48 & 0.22 & 0.22 & 0.22 & \colorbox{cyan}{\color{black}1} & 0.74 & 0.48 & 0.22 \\
			7 & 0.69 & \colorbox{cyan}{\color{black}1} & \colorbox{cyan}{\color{black}1} & 0.69 & 0.38 & \colorbox{cyan}{\color{black}1} & 0.69 & 0.38 & \colorbox{cyan}{\color{black}1} & 0.69 & 0.38 & \colorbox{cyan}{\color{black}1} & 0.69 & \colorbox{cyan}{\color{black}1} & 0.69 & \colorbox{cyan}{\color{black}1} & 0.69 & \colorbox{cyan}{\color{black}1} \\
			8 & 0.66 & \colorbox{cyan}{\color{black}1} & 0.66 & \colorbox{cyan}{\color{black}1} & 0.66 & \colorbox{cyan}{\color{black}1} & 0.66 & \colorbox{cyan}{\color{black}1} & \colorbox{cyan}{\color{black}1} & 0.66 & \colorbox{cyan}{\color{black}1} & \colorbox{cyan}{\color{black}1} & \colorbox{cyan}{\color{black}1} & 0.66 & \colorbox{cyan}{\color{black}1} & 0.66 & 0.32 & 0.32 \\
			9 & 0.65 & \colorbox{cyan}{\color{black}1} & 0.65 & \colorbox{cyan}{\color{black}1} & 0.65 & \colorbox{cyan}{\color{black}1} & 0.65 & \colorbox{cyan}{\color{black}1} & 0.65 & 0.65 & 0.30 & 0.30 & \colorbox{cyan}{\color{black}1} & 0.65 & 0.30 & 0.30 & \colorbox{cyan}{\color{black}1} & 0.65 \\
			10 & 0.70 & \colorbox{cyan}{\color{black}1} & 0.70 & 0.40 & \colorbox{cyan}{\color{black}1} & 0.70 & 0.40 & \colorbox{cyan}{\color{black}1} & 0.70 & \colorbox{cyan}{\color{black}1} & \colorbox{cyan}{\color{black}1} & 0.70 & 0.40 & \colorbox{cyan}{\color{black}1} & 0.70 & \colorbox{cyan}{\color{black}1} & 0.70 & 0.40 \\
			\bottomrule
	\end{tabular}}
\end{table*}

\subsubsection{Computational complexity of MANF-DNN-RP and MANF-RL-RP}

\begin{figure}[h]
	\vspace{-0.3cm} 
	\setlength{\abovecaptionskip}{0.2cm}   
	\setlength{\belowcaptionskip}{-0.3cm}   
	\subfigcapskip=-0.1cm 
	\centering
	\subfigure[loss value]{
		\begin{minipage}[t]{0.5\linewidth}
			\centering
			\includegraphics[width=4.2cm]{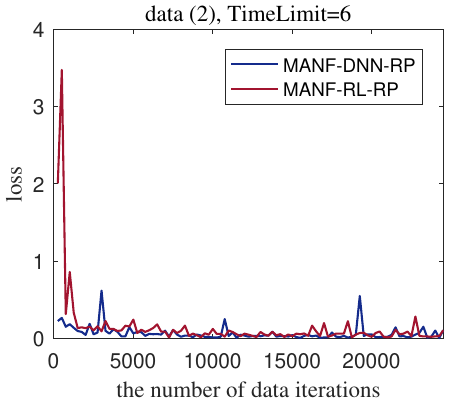}
			\label{figure101}
		\end{minipage}%
	}%
	\subfigure[task completion rate]{
		\begin{minipage}[t]{0.5\linewidth}
			\centering
			\includegraphics[width=4.3cm]{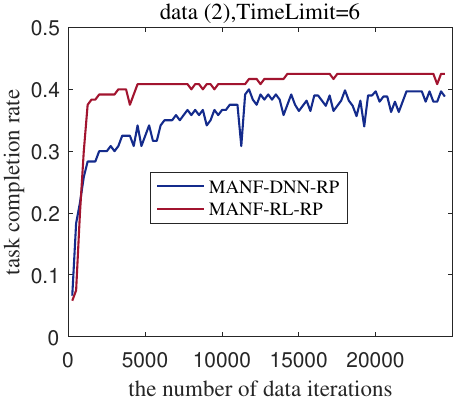}
			\label{figure102}
		\end{minipage}%
	}%
	\centering
	\caption{The variations of loss value and task completion rate as the number of data iterations increases.}
	\label{figure10}
\end{figure}

Based on data (2), we conducted a comprehensive analysis of the training process for MANF-DNN-RP and MANF-RL-RP. Figure \ref{figure10} depicts the variations in the loss value and task completion rate as the number of data iterations increases during model training. Based on Figure \ref{figure10}, we can observe two distinct characteristics in the curve. (1) Figure \ref{figure101} illustrates that within the first 4000 iterations, the training loss values of the two methods had already reached a lower level. However, as depicted in Figure \ref{figure102}, the convergence speed of the two models slows down after this point without reaching a convergence point. The reason behind this is that, at this stage, the two models have not fully completed the detection of the environmental space. They are exclusively trained using local detection results, which limits their ability to make optimal decisions in response to the overall environment. (2) Based on the observations from Figure \ref{figure102}, it is evident that MANF-RL-RP achieves faster convergence and demonstrates enhanced stability compared to MANF-DNN-RP. Specifically, MANF-DNN-RP exhibits higher levels of fluctuation, whereas MANF-RL-RP exhibits comparatively lower fluctuations once converged. However, in relation to \textbf{Algorithm \ref{algorithm1}} and \textbf{Algorithm \ref{algorithm2}}, MANF-DNN-RP and MANF-RL-RP have the same time complexity. What factors contribute to this discrepancy? The reason behind this is that, during the training process, the divergence in training data contributes to the observed disparities between the two models. Because the MANF-DNN-RP model primarily emphasizes immediate rewards, the variations in rewards per unit of time are not substantial. As a result, the distinguishability of training data labels becomes less evident, presenting a challenge for model fitting. In contrast,  MANF-RL-RP places emphasis on long-term decision-making benefits and demonstrates notable variations across different durations of the decision-making process. As a result, the distinguishability of training data labels becomes relatively prominent, facilitating smoother model fitting. Consequently, MANF-RL-RP achieves faster convergence and greater stability after convergence, in contrast to MANF-DNN-RP.


\subsubsection{Discussion on large-scale application scenarios}

In a typical crowd-sensing scenario, the number of UAVs, workers and cars may be large. When facing large-scale application scenarios, MANF-RL-RP needs to address some new issues. Therefore, the collaborative route planning of UAVs, workers and cars in large-scale application scenarios requires further research. The issues are as follows:

(1) Increased computational complexity: The increase of agents lead to a significantly higher computational complexity of the algorithm. Each agent needs to perform state sensing, decision-making, and learning, which may require larger computational resources and longer training times to handle a large-scale set of agents.

(2) Communication and coordination challenges: As the number of agents grows, communication and coordination among them become more difficult. It is necessary to design appropriate communication and collaboration mechanisms to ensure effective exchange of information and cooperation among agents for accomplishing sensing tasks.

(3) Increased complexity of environment interactions: In a multi-agent system, the interactions among agents add complexity to the environment. Agent's decision needs to consider the actions and strategies of other agents, while also being influenced by them. This raises challenges during training as agent needs to learn to adapt to the strategies and actions of other agents.

(4) Convergence and stability issues: With an increasing number of agents, convergence and stability problems may arise during the training process. Large-scale multi-agent systems can be more prone to getting trapped in local optima or exhibiting unstable action during learning. Appropriate algorithms and techniques need to be employed to address these challenges and ensure the system converges to good policies.

\section{Conclusion}
\label{section7}


Devastating disasters (e.g., earthquake) are extremely destructive. Efficiently obtaining the up-to-date information in the disaster-stricken area is the key to successful disaster response. UAVs, workers and cars can collaborate to complete the sensing tasks (e.g., data collection) in disaster-stricken areas. In this paper, we explicitly consider planning the routes of a group of agents (i.e., UAVs, workers, and cars) to maximize the task completion rate. We propose a heterogeneous multi-agent route planning algorithm MANF-RL-RP, which has the following design. (a) Global-local dual information processing. First, we mine the spatial features of global information based on convolutional neural networks (CNN) and share them with all agents to reduce the model training cost. Then, we divide the local information of agents into two parts: state information and filtering information. State information is used to guide the agents to make sequential decision. Filtering information is used to filter the non-optional actions to address the issue of sparse rewards in the sequential decision-marking process. (b) Model structure for heterogeneous multi-agent. We fill in the missing information of workers and cars to use the same data structure to represent the state of UAVs, workers, and cars, then share the same neural network parameter to reduce model parameter scale. Furthermore, we design a reasonable reward function and prove that UAVs, workers, and cars have cooperative relationships, which can guide model training well. In addition, we prove that the sequential decision-making process of agents has the Markov property, which simplifies the agent network structure. Finally, we conducted detailed experiments based on the rich simulation data. In comparison to the baseline algorithms, namely Greedy-SC-RP and MANF-DNN-RP, MANF-RL-RP has exhibited a significant improvement in terms of task completion rate. Under different $TimeLimit$, the task completion rate of MANF-RL-RP is significantly higher than that of MANF-DNN-RP and Greedy-SC-RP. When $TimeLimit$ is 6, the task completion rate is increased by 5.63\% and 42.86\% on average, respectively. When $TimeLimit$ is 9, the task completion rate is increased by 7.50\% and 56.94\% on average, respectively. When $TimeLimit$ is 12, the task completion rate is increased by 13.33\% and 70.60\% on average, respectively.

\bibliographystyle{IEEEtran}
\bibliography{refer}{}

\begin{thebibliography}{10}
\providecommand{\url}[1]{#1}
\csname url@samestyle\endcsname
\providecommand{\newblock}{\relax}
\providecommand{\bibinfo}[2]{#2}
\providecommand{\BIBentrySTDinterwordspacing}{\spaceskip=0pt\relax}
\providecommand{\BIBentryALTinterwordstretchfactor}{4}
\providecommand{\BIBentryALTinterwordspacing}{\spaceskip=\fontdimen2\font plus
\BIBentryALTinterwordstretchfactor\fontdimen3\font minus
  \fontdimen4\font\relax}
\providecommand{\BIBforeignlanguage}[2]{{%
\expandafter\ifx\csname l@#1\endcsname\relax
\typeout{** WARNING: IEEEtran.bst: No hyphenation pattern has been}%
\typeout{** loaded for the language `#1'. Using the pattern for}%
\typeout{** the default language instead.}%
\else
\language=\csname l@#1\endcsname
\fi
#2}}
\providecommand{\BIBdecl}{\relax}
\BIBdecl

\bibitem{gode72}
\url{https://en.wikipedia.org/wiki/Earthquake}, 2023.

\bibitem{2011Mobile}
R.~K. Ganti, Y.~Fan, and H.~Lei, ``Mobile crowdsensing: current state and
  future challenges,'' \emph{IEEE Communications Magazine}, vol.~49, no.~11,
  pp. 32--39, 2011.

\bibitem{2009Common}
P.~Dutta, P.~M. Aoki, N.~Kumar, A.~M. Mainwaring, and A.~Woodruff, ``Common
  sense: Participatory urban sensing using a network of handheld air quality
  monitors,'' in \emph{International Conference on Embedded Networked Sensor
  Systems}, 2009.

\bibitem{2011Discovery}
R.~Lee, S.~Wakamiya, and K.~Sumiya, ``Discovery of unusual regional social
  activities using geo-tagged microblogs,'' \emph{World Wide Web-internet \&
  Web Information Systems}, vol.~14, no.~4, pp. 321--349, 2011.

\bibitem{None2014How}
None, ``How long to wait? predicting bus arrival time with mobile phone based
  participatory sensing,'' \emph{IEEE Transactions on Mobile Computing},
  vol.~13, no.~6, pp. 1228--1241, 2014.

\bibitem{zhou2018mobile}
Z.~Zhou, J.~Feng, B.~Gu, B.~Ai, S.~Mumtaz, J.~Rodriguez, and M.~Guizani, ``When
  mobile crowd sensing meets uav: Energy-efficient task assignment and route
  planning,'' \emph{IEEE Transactions on Communications}, vol.~66, no.~11, pp.
  5526--5538, 2018.

\bibitem{liu2019energy}
C.~H. Liu, Z.~Chen, and Y.~Zhan, ``Energy-efficient distributed mobile crowd
  sensing: A deep learning approach,'' \emph{IEEE Journal on Selected Areas in
  Communications}, vol.~37, no.~6, pp. 1262--1276, 2019.

\bibitem{wang2021energy}
H.~Wang, C.~H. Liu, Z.~Dai, J.~Tang, and G.~Wang, ``Energy-efficient 3d
  vehicular crowdsourcing for disaster response by distributed deep
  reinforcement learning,'' in \emph{Proceedings of the 27th ACM SIGKDD
  Conference on Knowledge Discovery \& Data Mining}, 2021, pp. 3679--3687.

\bibitem{liu2020curiosity}
C.~H. Liu, Y.~Zhao, Z.~Dai, Y.~Yuan, G.~Wang, D.~Wu, and K.~K. Leung,
  ``Curiosity-driven energy-efficient worker scheduling in vehicular
  crowdsourcing: A deep reinforcement learning approach,'' in \emph{2020 IEEE
  36th International Conference on Data Engineering (ICDE)}.\hskip 1em plus
  0.5em minus 0.4em\relax IEEE, 2020, pp. 25--36.

\bibitem{dai2022aoi}
Z.~Dai, C.~H. Liu, Y.~Ye, R.~Han, Y.~Yuan, G.~Wang, and J.~Tang, ``Aoi-minimal
  uav crowdsensing by model-based graph convolutional reinforcement learning,''
  in \emph{IEEE INFOCOM 2022-IEEE Conference on Computer Communications}.\hskip
  1em plus 0.5em minus 0.4em\relax IEEE, 2022, pp. 1029--1038.

\bibitem{liu2020energy}
C.~H. Liu, C.~Piao, and J.~Tang, ``Energy-efficient uav crowdsensing with
  multiple charging stations by deep learning,'' in \emph{IEEE INFOCOM
  2020-IEEE Conference on Computer Communications}.\hskip 1em plus 0.5em minus
  0.4em\relax IEEE, 2020, pp. 199--208.

\bibitem{liu2019distributed}
C.~H. Liu, X.~Ma, X.~Gao, and J.~Tang, ``Distributed energy-efficient multi-uav
  navigation for long-term communication coverage by deep reinforcement
  learning,'' \emph{IEEE Transactions on Mobile Computing}, vol.~19, no.~6, pp.
  1274--1285, 2019.

\bibitem{silver2017mastering}
D.~Silver, J.~Schrittwieser, K.~Simonyan, I.~Antonoglou, A.~Huang, A.~Guez,
  T.~Hubert, L.~Baker, M.~Lai, A.~Bolton \emph{et~al.}, ``Mastering the game of
  go without human knowledge,'' \emph{nature}, vol. 550, no. 7676, pp.
  354--359, 2017.

\bibitem{mnih2015human}
V.~Mnih, K.~Kavukcuoglu, D.~Silver, A.~A. Rusu, J.~Veness, M.~G. Bellemare,
  A.~Graves, M.~Riedmiller, A.~K. Fidjeland, G.~Ostrovski \emph{et~al.},
  ``Human-level control through deep reinforcement learning,'' \emph{nature},
  vol. 518, no. 7540, pp. 529--533, 2015.

\bibitem{zhao2016spatial}
Y.~Zhao and Q.~Han, ``Spatial crowdsourcing: current state and future
  directions,'' \emph{IEEE communications magazine}, vol.~54, no.~7, pp.
  102--107, 2016.

\bibitem{tong2016online}
Y.~Tong, J.~She, B.~Ding, L.~Wang, and L.~Chen, ``Online mobile micro-task
  allocation in spatial crowdsourcing,'' in \emph{2016 IEEE 32Nd international
  conference on data engineering (ICDE)}.\hskip 1em plus 0.5em minus
  0.4em\relax IEEE, 2016, pp. 49--60.

\bibitem{li2021simultaneous}
B.~Li, Y.~Cheng, Y.~Yuan, G.~Wang, and L.~Chen, ``Simultaneous arrival matching
  for new spatial crowdsourcing platforms,'' in \emph{Proceedings of the
  Twenty-Ninth International Conference on International Joint Conferences on
  Artificial Intelligence}, 2021, pp. 1279--1287.

\bibitem{li2019three}
------, ``Three-dimensional stable matching problem for spatial crowdsourcing
  platforms,'' in \emph{Proceedings of the 25th ACM SIGKDD International
  Conference on Knowledge Discovery \& Data Mining}, 2019, pp. 1643--1653.

\bibitem{codeanddata}
\url{https://github.com/CocaColaZero/MANF-for-Collaborative-Route-Planning.git},
  2023.

\bibitem{reddy2009using}
S.~Reddy, K.~Shilton, J.~Burke, D.~Estrin, M.~Hansen, and M.~Srivastava,
  ``Using context annotated mobility profiles to recruit data collectors in
  participatory sensing,'' in \emph{International Symposium on Location-and
  Context-Awareness}.\hskip 1em plus 0.5em minus 0.4em\relax Springer, 2009,
  pp. 52--69.

\bibitem{zhang2014crowdrecruiter}
D.~Zhang, H.~Xiong, L.~Wang, and G.~Chen, ``Crowdrecruiter: Selecting
  participants for piggyback crowdsensing under probabilistic coverage
  constraint,'' in \emph{Proceedings of the 2014 ACM International Joint
  Conference on Pervasive and Ubiquitous Computing}, 2014, pp. 703--714.

\bibitem{xiong2015icrowd}
H.~Xiong, D.~Zhang, G.~Chen, L.~Wang, V.~Gauthier, and L.~E. Barnes, ``icrowd:
  Near-optimal task allocation for piggyback crowdsensing,'' \emph{IEEE
  Transactions on Mobile Computing}, vol.~15, no.~8, pp. 2010--2022, 2015.

\bibitem{song2014qoi}
Z.~Song, B.~Zhang, C.~H. Liu, A.~V. Vasilakos, J.~Ma, and W.~Wang, ``Qoi-aware
  energy-efficient participant selection,'' in \emph{2014 Eleventh Annual IEEE
  International Conference on sensing, communication, and networking
  (SECON)}.\hskip 1em plus 0.5em minus 0.4em\relax IEEE, 2014, pp. 248--256.

\bibitem{karaliopoulos2015user}
M.~Karaliopoulos, O.~Telelis, and I.~Koutsopoulos, ``User recruitment for
  mobile crowdsensing over opportunistic networks,'' in \emph{2015 IEEE
  Conference on Computer Communications (INFOCOM)}.\hskip 1em plus 0.5em minus
  0.4em\relax IEEE, 2015, pp. 2254--2262.

\bibitem{yu2018participant}
Z.~Yu, J.~Zhou, W.~Guo, L.~Guo, and Z.~Yu, ``Participant selection for t-sweep
  k-coverage crowd sensing tasks,'' \emph{World Wide Web}, vol.~21, no.~3, pp.
  741--758, 2018.

\bibitem{wang2017multi}
L.~Wang, Z.~Yu, Q.~Han, B.~Guo, and H.~Xiong, ``Multi-objective optimization
  based allocation of heterogeneous spatial crowdsourcing tasks,'' \emph{IEEE
  Transactions on Mobile Computing}, vol.~17, no.~7, pp. 1637--1650, 2017.

\bibitem{zhang2015quality}
M.~Zhang, P.~Yang, C.~Tian, S.~Tang, X.~Gao, B.~Wang, and F.~Xiao,
  ``Quality-aware sensing coverage in budget-constrained mobile crowdsensing
  networks,'' \emph{IEEE Transactions on Vehicular Technology}, vol.~65, no.~9,
  pp. 7698--7707, 2015.

\bibitem{wang2018multi}
J.~Wang, Y.~Wang, D.~Zhang, F.~Wang, H.~Xiong, C.~Chen, Q.~Lv, and Z.~Qiu,
  ``Multi-task allocation in mobile crowd sensing with individual task quality
  assurance,'' \emph{IEEE Transactions on Mobile Computing}, vol.~17, no.~9,
  pp. 2101--2113, 2018.

\bibitem{li2015dynamic}
H.~Li, T.~Li, and Y.~Wang, ``Dynamic participant recruitment of mobile crowd
  sensing for heterogeneous sensing tasks,'' in \emph{2015 IEEE 12th
  International Conference on Mobile Ad Hoc and Sensor Systems}.\hskip 1em plus
  0.5em minus 0.4em\relax IEEE, 2015, pp. 136--144.

\bibitem{wang2018heterogeneous}
L.~Wang, Z.~Yu, D.~Zhang, B.~Guo, and C.~H. Liu, ``Heterogeneous multi-task
  assignment in mobile crowdsensing using spatiotemporal correlation,''
  \emph{IEEE Transactions on Mobile Computing}, vol.~18, no.~1, pp. 84--97,
  2018.

\bibitem{wang2019user}
E.~Wang, Y.~Yang, and K.~Lou, ``User selection utilizing data properties in
  mobile crowdsensing,'' \emph{Information Sciences}, vol. 490, pp. 210--226,
  2019.

\bibitem{han2021online}
L.~Han, Z.~Yu, Z.~Yu, L.~Wang, H.~Yin, and B.~Guo, ``Online organizing
  large-scale heterogeneous tasks and multi-skilled participants in mobile
  crowdsensing,'' \emph{IEEE Transactions on Mobile Computing}, 2021.

\bibitem{2016TaskMe}
L.~Yan, B.~Guo, W.~Yang, W.~Wu, and D.~Zhang, ``Taskme: multi-task allocation
  in mobile crowd sensing,'' \emph{ACM}, 2016.

\bibitem{2017Joint}
T.~X. Tran and D.~Pompili, ``Joint task offloading and resource allocation for
  multi-server mobile-edge computing networks,'' \emph{IEEE Transactions on
  Vehicular Technology}, 2017.

\bibitem{2017Multiuser}
X.~Lyu, T.~Hui, C.~Sengul, and Z.~Ping, ``Multiuser joint task offloading and
  resource optimization in proximate clouds,'' \emph{IEEE Transactions on
  Vehicular Technology}, vol.~66, no.~4, pp. 1--1, 2017.

\bibitem{lowe2017multi}
R.~Lowe, Y.~I. Wu, A.~Tamar, J.~Harb, O.~Pieter~Abbeel, and I.~Mordatch,
  ``Multi-agent actor-critic for mixed cooperative-competitive environments,''
  \emph{Advances in neural information processing systems}, vol.~30, 2017.

\bibitem{foerster2018counterfactual}
J.~Foerster, G.~Farquhar, T.~Afouras, N.~Nardelli, and S.~Whiteson,
  ``Counterfactual multi-agent policy gradients,'' in \emph{Proceedings of the
  AAAI conference on artificial intelligence}, vol.~32, no.~1, 2018.

\bibitem{sunehag2017value}
P.~Sunehag, G.~Lever, A.~Gruslys, W.~M. Czarnecki, V.~Zambaldi, M.~Jaderberg,
  M.~Lanctot, N.~Sonnerat, J.~Z. Leibo, K.~Tuyls \emph{et~al.},
  ``Value-decomposition networks for cooperative multi-agent learning,''
  \emph{arXiv preprint arXiv:1706.05296}, 2017.

\bibitem{rashid2018qmix}
T.~Rashid, M.~Samvelyan, C.~Schroeder, G.~Farquhar, J.~Foerster, and
  S.~Whiteson, ``Qmix: Monotonic value function factorisation for deep
  multi-agent reinforcement learning,'' in \emph{International conference on
  machine learning}.\hskip 1em plus 0.5em minus 0.4em\relax PMLR, 2018, pp.
  4295--4304.

\bibitem{rashid2020weighted}
T.~Rashid, G.~Farquhar, B.~Peng, and S.~Whiteson, ``Weighted qmix: Expanding
  monotonic value function factorisation for deep multi-agent reinforcement
  learning,'' \emph{Advances in neural information processing systems},
  vol.~33, pp. 10\,199--10\,210, 2020.

\bibitem{son2019qtran}
K.~Son, D.~Kim, W.~J. Kang, D.~E. Hostallero, and Y.~Yi, ``Qtran: Learning to
  factorize with transformation for cooperative multi-agent reinforcement
  learning,'' in \emph{International conference on machine learning}.\hskip 1em
  plus 0.5em minus 0.4em\relax PMLR, 2019, pp. 5887--5896.

\bibitem{liu2020multi}
C.~H. Liu, Z.~Dai, H.~Yang, and J.~Tang, ``Multi-task-oriented vehicular
  crowdsensing: A deep learning approach,'' in \emph{IEEE INFOCOM 2020-IEEE
  Conference on Computer Communications}.\hskip 1em plus 0.5em minus
  0.4em\relax IEEE, 2020, pp. 1123--1132.

\bibitem{nemhauser1978analysis}
G.~L. Nemhauser, L.~A. Wolsey, and M.~L. Fisher, ``An analysis of
  approximations for maximizing submodular set functions—i,''
  \emph{Mathematical programming}, vol.~14, no.~1, pp. 265--294, 1978.

\bibitem{hare2019dealing}
J.~Hare, ``Dealing with sparse rewards in reinforcement learning,'' \emph{arXiv
  preprint arXiv:1910.09281}, 2019.

\bibitem{ha2016hypernetworks}
D.~Ha, A.~Dai, and Q.~V. Le, ``Hypernetworks,'' \emph{arXiv preprint
  arXiv:1609.09106}, 2016.

\bibitem{9641503}
L.~Han, Z.~Yu, Z.~Yu, L.~Wang, H.~Yin, and B.~Guo, ``Online organizing
  large-scale heterogeneous tasks and multi-skilled participants in mobile
  crowdsensing,'' \emph{IEEE Transactions on Mobile Computing}, pp. 1--1, 2021.

\bibitem{2011Friendship}
E.~Cho, S.~A. Myers, and J.~Leskovec, ``Friendship and mobility: user movement
  in location-based social networks,'' in \emph{Proceedings of the 17th ACM
  SIGKDD International Conference on Knowledge Discovery and Data Mining, San
  Diego, CA, USA, August 21-24, 2011}, 2011.

\bibitem{1989Approximation}
G.~Cybenko, ``Approximation by superpositions of a sigmoidal function,''
  \emph{Mathematics of Control, Signals and Systems}, vol.~2, no.~4, pp.
  303--314, 1989.

\bibitem{wang2022task}
L.~Wang, D.~Yang, Z.~Yu, F.~Xiong, L.~Han, S.~Pan, and B.~Guo, ``Task
  scheduling in three-dimensional spatial crowdsourcing: A social welfare
  perspective,'' \emph{IEEE Transactions on Mobile Computing}, 2022.

\end{thebibliography}

\end{document}